\RequirePackage{fix-cm}
\documentclass[twocolumn]{svjour3}          
\smartqed  

\usepackage{amsmath}
\usepackage{gensymb}
\usepackage{diagbox}
\usepackage{graphics}
\usepackage{epsfig}
\usepackage{threeparttable}
\usepackage{xspace}
\usepackage{siunitx}
\usepackage{graphicx}
\usepackage{subfig}
\usepackage{placeins}
\usepackage{blindtext}
\usepackage{amssymb}
\usepackage{pifont}
\usepackage{threeparttable}
\usepackage{color}
\usepackage[normalem]{ulem}
\usepackage{multirow}
\usepackage{float}
\usepackage{amsfonts}
\usepackage{bm}
\usepackage{bbm}
\usepackage{array}
\usepackage{tabulary}	
\usepackage{microtype}
\usepackage{booktabs}
\usepackage{xspace}
\usepackage[table]{xcolor}
\usepackage{colortbl}
\usepackage{diagbox}
\usepackage{rotating}
\usepackage{booktabs}
\usepackage{overpic}
\usepackage{enumitem}
\usepackage{colortbl}
\usepackage{soul}
\usepackage{url}
\usepackage{makecell, verbatim, sidecap}
\usepackage{cite}
\usepackage[T1]{fontenc}
\usepackage{bbm}
\usepackage{tikz}
\usepackage{comment}
\usepackage{graphicx}
\usepackage{amssymb}
\usepackage{booktabs}
\usepackage{algorithm}
\usepackage{algpseudocode}
\usepackage[accsupp]{axessibility} 
\usepackage{xfrac}
\usepackage{color}
\usepackage{wrapfig}
\usepackage{microtype}
\usepackage{tabularx}
\usepackage{booktabs}
\usepackage{multirow}
\usepackage{blindtext}
\usepackage{xspace}

\definecolor{citecolor}{HTML}{0071bc}
\usepackage[pagebackref=true,breaklinks=true,colorlinks,citecolor=citecolor,bookmarks=false]{hyperref}

\definecolor{scorered}{HTML}{e4485a}
\definecolor{scoreblue}{HTML}{4a7ee8}
\definecolor{scoregreen}{HTML}{80ba0e}
\definecolor{scoreyellow}{HTML}{d8ac0d}
\definecolor{scorepurple}{HTML}{846bc8}

\definecolor{purple0}{HTML}{e9e9f3}
\definecolor{purple}{HTML}{dcdaed}
\definecolor{purple1}{HTML}{bab6da}
\definecolor{blue0}{HTML}{b6d9f0}
\definecolor{blue1}{HTML}{80b1d1}
\definecolor{blue2}{HTML}{2a8ed1}
\definecolor{blue3}{HTML}{0071bc}

\definecolor{mygray00}{gray}{.3}
\definecolor{mygray0}{gray}{.6}
\definecolor{mygray}{gray}{.85}
\definecolor{mygray1}{gray}{.9}
\definecolor{mygray2}{gray}{.95}

\makeatletter
\newcommand{\thickhline}{%
    \noalign {\ifnum 0=`}\fi \hrule height 1pt
    \futurelet \reserved@a \@xhline
}
\makeatother

\makeatletter
\DeclareRobustCommand\onedot{\futurelet\@let@token\@onedot}
\def\@onedot{\ifx\@let@token.\else.\null\fi\xspace}

\makeatother

\newcommand{\app}{\raise.17ex\hbox{$\scriptstyle\sim$}}

\newcommand{\cmark}{\ding{51}}%
\newcommand{\xmark}{\ding{55}}%

\journalname{IJCV}
\begin{document}
\title{A Survey on Human Interaction Motion Generation}
\sloppy


\author{Kewei Sui $^{1}$ \and
        Anindita Ghosh* $^{2}$ \and
        Inwoo Hwang* $^{3}$ \and
        Bing Zhou $^1$ \and \\
        Jian Wang $^1$ \and
        Chuan Guo $^1$ 
}

\institute{Kewei Sui \at
              \email{ksui@snapchat.com}
           \and
            Anindita Ghosh\at
              \email{anindita.ghosh@dfki.de}
           \and
           Inwoo Hwang \at
              \email{inwoohwang0818@gmail.com}
           \and
           Bing Zhou \at
              \email{bzhou@snapchat.com}
           \and
           Jian Wang \at
              \email{jwang4@snapchat.com}
           \and
           Chuan Guo \at
              \email{cguo2@snapchat.com}
           \\
           \\
           $^1$ Snap Inc., California, USA
           \\
           $^2$ Saarland Informatics Campus, DFKI, MPI Informatics, Germany
           \\
           $^3$ Seoul National University, Seoul, South Korea
           \\
           * Indicates equal contribution
           \\
           (Corresponding author: Chuan Guo)
}

\date{Received: xx Mar 2025 / Accepted: xx xx 2025}

\maketitle

\begin{abstract}
Humans inhabit a world defined by interactions—with other humans, objects, and environments. These interactive movements not only convey our relationships with our surroundings but also demonstrate how we perceive and communicate with the real world. Therefore, replicating these interaction behaviors in digital systems has emerged as an important topic for applications in robotics, virtual reality, and animation. While recent advances in deep generative models and new datasets have accelerated progress in this field, significant challenges remain in modeling the intricate human dynamics and their interactions with entities in the external world. In this survey, we present, \textit{for the first time}, a comprehensive overview of the literature in human interaction motion generation. We begin by establishing foundational concepts essential for understanding the research background. We then systematically review existing solutions and datasets across three primary interaction tasks—human-human, human-object, and human-scene interactions—followed by evaluation metrics. Finally, we discuss open research directions and future opportunities. The repository listing relevant papers is accessible at: \url{https://github.com/soraproducer/Awesome-Human-Interaction-Motion-Generation}.

\end{abstract}

\section{Introduction}
\label{sec:introduction}

\begin{figure*}[h]
	\begin{center}
		\includegraphics[width=0.95\linewidth]{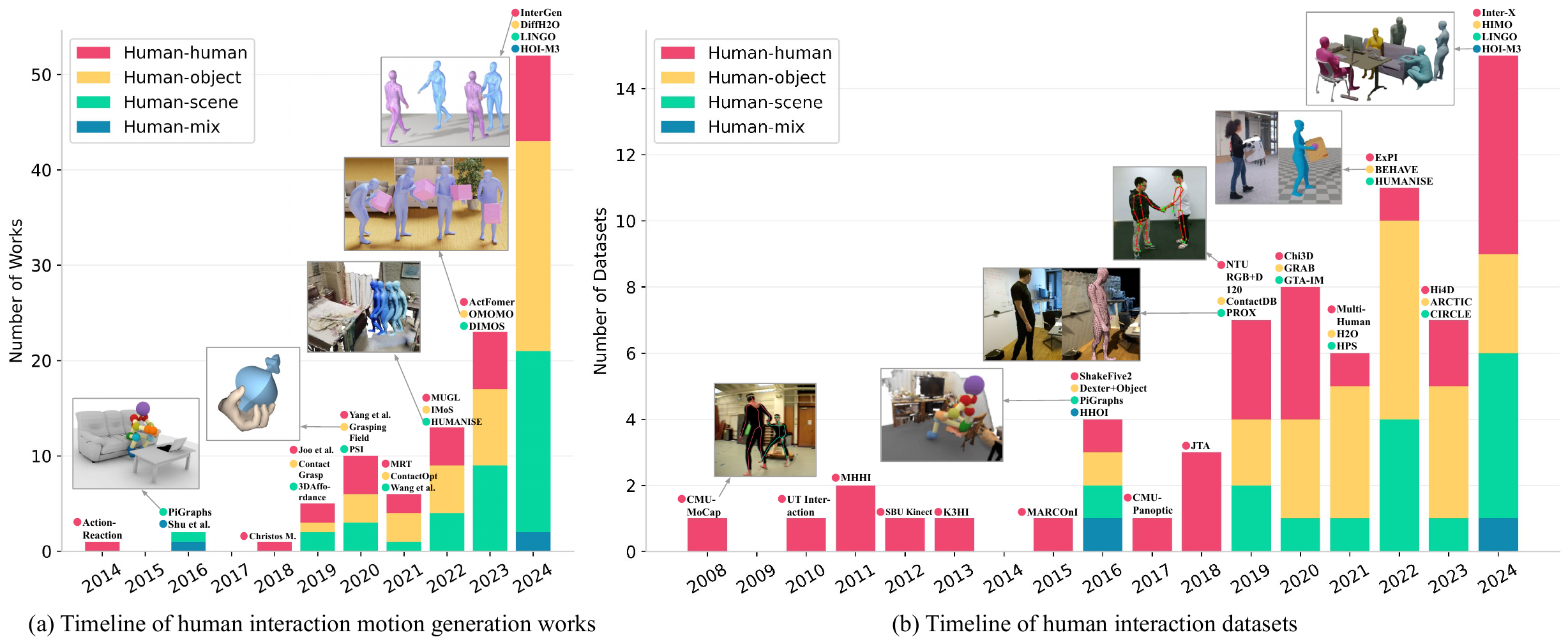}
	\end{center}
	\vspace{-10pt}
	\captionsetup{font=small}
	\caption{Statistics on the number of works and datasets on human interaction motion generation over the past two decades, categorized into four interaction scenarios.}
        \label{fig:timeline}
	\vspace{-10pt}
\end{figure*}

Human life is fundamentally characterized by interactions with the external environment through motion~\cite{clark1998being, norman1995psychopathology}. These interactions range from everyday actions, such as using a smartphone and cooking, to social gestures, such as handshakes and clapping. Successfully understanding and reproducing such complex behaviors is crucial for developing human-like entities across various domains, including 3D virtual characters in entertainment media~\cite{arikan2003motion, de2018applying}, humanoid robots~\cite{naksuk2005whole, kim2009stable}, and digital avatars~\cite{badler1999animation, van2010real, kuo2012motion}.

The past decade has witnessed remarkable progress in generative modeling across multiple domains: text~\cite{brown2020language, raffel2020exploring, touvron2023llama}, images~\cite{karras2019style, ramesh2021zero, rombach2022high}, video~\cite{vondrick2016generating, tulyakov2018mocogan, singer2022make}, audio~\cite{van2016wavenet, huang2018music, dhariwal2020jukebox}, and 3D objects~\cite{wu2016learning, park2019deepsdf, mildenhall2021nerf}. This advancement has been driven by foundational generative models, including Variational Autoencoders (VAEs)\cite{kingma2013vae}, Generative Adversarial Networks (GANs)\cite{gui2021ganreview}, Diffusion models~\cite{yang2023diffusion, croitoru2023diffusion}, Large Language Models (LLMs)~\cite{minaee2024llmsurvey}, and Vision-Language Models (VLMs)~\cite{li2025visual}. These developments have also enhanced our ability to generate diverse and natural 3D human motions from various inputs such as action categories\cite{guo2020action2motion,lucas2022posegpt,petrovich2021action}, textual descriptions~\cite{guo2022generating,guidedmotiondiffusion,petrovich2022temos,tevet2023human,zhang2023generating,chen2023executing}, audio~\cite{tseng2023edge,alexanderson2023listen,gong2023tm2d,siyao2022bailando,li2021ai}, and so on.

However, generating human \textit{interaction} motions presents distinct challenges beyond standard generative modeling approaches. First, human interaction is inherently stochastic, yet the resulting body movements must maintain spatial and temporal coherence that aligns with specific human intentions. Second, interacting with the external world demands environmental awareness, requiring adaptation to diverse scene layouts, understanding of object properties and affordances, and compliance with physical constraints to prevent intra- and inter-penetration. Last but not least, the collection of human interaction data is resource-intensive and difficult to scale, making it impractical to rely solely on data-driven learning. Therefore, incorporating domain expertise into learning models is essential to complement traditional generative methods. In summary, generating natural human interaction motions requires the ability to model human dynamics, incorporate physical constraints, and understand the spatial semantics and relationships within the holistic environment.

Despite these challenges, research on human interaction generation has advanced rapidly in the last decade, with growing interest over time. Fig.~\ref{fig:timeline} chronicles these developments, highlighting key milestones covered in this survey. We categorize human interaction scenarios in existing motions into four main types: human-human interaction (HHI), human-object interaction (HOI), human-scene interaction (HSI), and human-mix interaction (involving multiple interaction types simultaneously).  This survey provides a comprehensive review of interactive human motion generation, addressing recent advances and emerging challenges. The paper is structured as follows. In Section \S\ref{sec:scope}, we define the scope of this survey and identify related topics beyond its scope. Section \S\ref{sec:preliminaries} covers the preliminaries, providing foundational knowledge and key concepts essential for understanding the subsequent sections. Section \S\ref{sec:interactive-hmg} reviews the various methods and techniques employed in interactive human motion generation. In Section \S\ref{sec:datasets}, we provide an overview of the commonly used datasets in this field, highlighting their distinct features. Section \S\ref{sec:eval} explores the evaluation metrics utilized to measure the performance of these methods. Finally, Section \S\ref{sec:conclusion} summarizes the current landscape and offers an exploration of future research directions. This survey aims to provide researchers and practitioners with a comprehensive understanding of the state of the art in this rapidly evolving field.

\section{Scope}
\label{sec:scope}

\begin{figure*}[h]
	\begin{center}
		\includegraphics[width=0.95\linewidth]{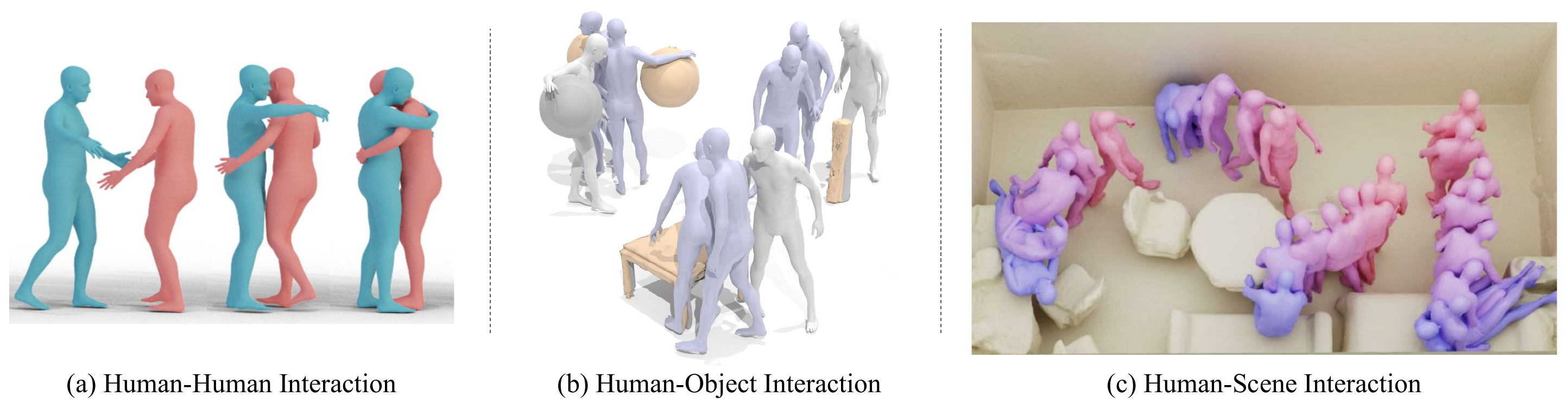}
	\end{center}
	\vspace{-10pt}
	\captionsetup{font=small}
	\caption{Illustration of three major human interaction motion generation tasks: (a) Human-human Interaction; (b) Human-object Interaction; and (c) Human-scene Interaction. Figures are adapted from~\cite{xu2023interx,xu2023interdiff,TRUMANS}.}
	\label{fig:interaction_viz}
	\vspace{-10pt}
\end{figure*}

This survey examines interactive human motion generation, with a focus on generation methods, datasets, and evaluation metrics across four key interaction types illustrated in Fig.~\ref{fig:interaction_viz}: human-human, human-object, human-scene, and human-mix interactions. Our investigation encompasses various generation approaches, including interactive motion generation, motion prediction, and physics-based simulation. The scope of this survey specifically excludes human motion tasks that do not involve generation or interactions. Related but distinct research areas include single-person motion generation~\cite{zhu2023hmgsurvey}, human motion style transfer~\cite{sensors23052597}, human pose estimation~\cite{Chen_2020, liu2021recentadvancesmonocular2d}, and human action recognition~\cite{kong2022harsurvey}. For comprehensive reviews of these topics, the readers can refer to the respective surveys cited above.

\section{Preliminaries}
\label{sec:preliminaries}
\vspace{-6pt}

This section establishes the fundamental concepts of human interaction motion generation. We examine three key aspects: the entities involved in interactions, the conditions governing interaction motions, and the core methodologies for generating these motions. This foundation provides essential context for understanding the research developments discussed throughout the survey.

\subsection{Interactive Entities}

\subsubsection{Human Motion}

Human motion is a fundamental component of interactions. Accurate motion capture and efficient motion representation are essential for human interaction motion generation models.

\paragraph{Human Motion Capture.} Human movements can be captured through several approaches, each with distinct trade-offs. Marker-based optical systems (e.g., Vicon~\cite{merriaux2017study} and OptiTrack~\cite{furtado2019comparative}) track markers attached to key joints using multiple optical cameras, providing the highest precision but at a significant cost. Inertial-based motion capture systems offer an affordable alternative using IMU sensors or Smartsuits~\cite{mihcin2019investigation} to track body segment movements, although they require regular calibration to address sensor drift. RGB-D cameras (e.g., Kinect~\cite{zhang2012microsoft}) enable low-cost motion capture through single or multi-view setups, extracting 3D joint information from RGB and depth data, but typically lack fine motion details. Recent deep learning-based pose estimation methods~\cite{kocabas2020vibe,wang2024tram} can reconstruct 3D motions from video footage, although their generalization capabilities remain limited. Additionally, 3D graphics engines provide a flexible option for generating synthetic human motions in a virtual 3D environment, without physical capture equipment.

\paragraph{Representation.} In kinematic-based methods, human motion is represented as a sequence of skeletal poses defined by joints or bones in 3D space. These motions can be expressed through either 3D joint positions or bone rotations along kinematic chains (e.g., limbs and spine). Recent studies~\cite{villegas2018neural, zhou2019continuity} favor rotation-based representations as they inherently encode skeletal topology. While traditional rotation formats (Euler angles and quaternions) are available, the 6D rotation representation~\cite{zhou2019continuity} has gained prominence for its continuity and compatibility with deep learning models. Parametric models such as SMPL~\cite{smpl}, SMPL-X~\cite{pavlakos2019smpl-x}, and GHUM~\cite{Xu_2020_CVPR_GHUM} extend beyond rotational pose parameters by incorporating shape parameters. These models parameterize surface vertices and deformations using both pose and shape information, enabling geometry-aware motion representation, which is crucial for fine-grained interactions.

\subsubsection{Object}

Objects also serve as important interactive entities. Capturing their shape and motion accurately, along with structured representation, is essential for object-centric interaction modeling.

\paragraph{Object Shape and Motion Capture.} To capture both the shape and motion of 3D objects, several complementary approaches are typically used. For shape capture, methods include 3D scanning (using laser scanners~\cite{ebrahim20153d}, structured light~\cite{zhang2018high}, or photogrammetry~\cite{surmen2023photogrammetry}), RGB-D sensors~\cite{zollhofer2018state}, or multi-view stereo reconstruction~\cite{seitz2006comparison}. Object motion can be tracked either using marker-based approaches, where reflective markers are attached to key points and tracked with optical motion capture systems, or through markerless methods that rely on visual feature tracking~\cite{shi1994good}, dense surface tracking~\cite{newcombe2011kinectfusion}, or deep learning-based pose estimation~\cite{xiang2017posecnn}.
Articulated objects require additional consideration. The first step typically involves modeling the object's kinematic structure, defining joint hierarchies, and establishing articulation constraints. Capture typically involves tracking individual rigid segments using markers placed at joints and key points, followed by articulation reconstruction from marker trajectories~\cite{jiang2023full}. To ensure realistic dynamics, these captured motions are often enhanced with physics-based simulations that model collision responses, gravitational effects, and inertial forces~\cite{wang2023deepsimho, wang2023physhoi, zhang2024force}.

\paragraph{Representation.}
In human-object interaction synthesis, objects are represented in various formats to capture their geometric and physical attributes for deep learning models. Point clouds capture object surfaces as unordered 3D points, offering flexibility for complex shape modeling~\cite{guo2020deep}. Meshes provide high-resolution geometric details through vertices and edges, enabling precise contact modeling. Basis Point Sets (BPS)~\cite{prokudin2019efficient} encode object geometry in fixed-dimensional representations, balancing efficient neural processing with robustness to shape variations.
For rigid objects in motion, the 6 Degrees of Freedom (6DoF) format represents both translation and orientation as $\mathbf{T}{1:N} = [\mathbf{t}, \mathbf{R}]{1:N}$, where $N$ is the total frame count, $\mathbf{t} \in \mathbb{R}^3$ denotes translation, and $\mathbf{R} \in SO(3)$ represents rotation at each frame.
Articulated objects~\cite{fan2023arctic} are modeled using a 3D mesh $\mathcal{O}(\Omega) \in \mathbb{R}^{V \times 3}$, where $V$ represents the count of vertices on the object surface. The object pose $\Omega \in \mathbb{R}^7$ combines articulation rotation $(\mathbf{\omega} \in \mathbb{R})$, object translation $\mathbf{t} \in \mathbb{R}^3$, and object rotation $\mathbf{R} \in SO(3)$.
These representations are often enhanced with object-centric interaction regions and canonicalization techniques to standardize spatial relationships.

\subsubsection{Scene}

Scenes provide the spatial and contextual foundation for interactions, necessitating accurate acquisition methods and structured representations to model human-scene relationships effectively.

\paragraph{Scene Acquisition.}
Human-scene interactions can be captured in both real and virtual environments. Real-world scenes are digitized using advanced scanning technologies: LiDAR systems capture high-resolution 3D point clouds~\cite{cong2024laserhuman}, while structured light scanning~\cite{PROX, kim2024parahome} reconstructs detailed surface geometry. Alternatively, existing 3D scene datasets, such as ScanNet~\cite{dai2017scannet}, provide ready-to-use virtual environments~\cite{humanise}.
Recent approaches have expanded scene diversity through synthetic virtual environments~\cite{LINGO, TRUMANS, CIRCLE, caoHMP2020}. Created using 3D modeling tools like Unity~\cite{puig2018virtualhome}, Unreal Engine~\cite{lin2016virtual}, or Blender~\cite{denninger2020blenderproc}, these environments offer precise control over scene parameters, including textures, lighting, and object placement. This approach enables the scalable capture of complex human-scene interactions across diverse settings.

\paragraph{Representation.}
Point clouds are widely used due to their lightweight nature and their ability to preserve detailed geometric information. Each point contains the 3D spatial coordinates of scene surfaces, with additional features like surface normals or semantic labels. These representations are typically processed using specialized architectures, like PointNet\cite{qi2017pointnet} or PointTransformer\cite{zhao2021point}.
Occupancy grids and voxel representations discretize 3D space into regular cells containing binary information. These approaches facilitate efficient collision checking and spatial reasoning, making them particularly valuable for human-scene interaction tasks. Various architectures, including 3D Convolutional Networks (3D-CNNs)~\cite{tran2015learning} and Vision Transformers (ViTs)~\cite{dosovitskiy2021imageworth16x16words}, have been employed to process these representations. Similar to object representation, BPS~\cite{prokudin2019efficient} features also offer a structured encoding of scene geometry by measuring point-wise distances to a predefined set of basis points.

\subsection{Conditioning Modalities}

Human interaction motion synthesis often conditions on other modalities. These modalities provide additional context or constraints, enabling more controllable and semantically consistent motion generation.

\textbf{Text} Textual descriptions have emerged as a popular modality for guiding human interaction motion generation~\cite{shan2024opendomain, fan2024freemotion, ponce2024in2in, wang2024intercontrol, shafir2023ComMDM, milacski2024ghostgroundedhumanmotion, InterGen, yang2024f}. Text-based guidance enables models to process detailed instructions that define interactions between generated human motions and various entities (humans, objects, scenes, or combinations thereof). These text conditions are typically incorporated either as embeddings—such as CLIP~\cite{radford2021learning} embeddings~\cite{shan2024opendomain, fan2024freemotion, ponce2024in2in, wang2024intercontrol, shafir2023ComMDM, milacski2024ghostgroundedhumanmotion}—as penultimate layer outputs from LLMs\cite{InterGen}, or as sequences of discrete word tokens\cite{yang2024f}.

\textbf{Audio} Audio-driven approaches enable models to generate interaction motions synchronized with acoustic cues, typically in HHI scenarios~\cite{ahuja2019reactornot, Yang2020, siyao2024duolando}. The audio input manifests either as conversational exchanges between actors and reactors~\cite{ahuja2019reactornot, Yang2020} or as musical accompaniment~\cite{siyao2024duolando} coordinating multiple participants. These acoustic signals are processed into salient features—including prosody, excitation, music intensity, and rhythmic beats—using established tools such as OpenSmile~\cite{eyben2013recent} and Librosa~\cite{mcfee2015librosa}.

\textbf{Action Class} Action classes serve as a well-established conditioning mechanism in huaman interaction motion generation~\cite{goel2022interactionmixmatch, maheshwari2021mugl, Gupta_2023_DSAG, xu2022actformer}. These categorical descriptors are typically implemented as one-hot encodings~\cite{goel2022interactionmixmatch, maheshwari2021mugl, Gupta_2023_DSAG} or label token embeddings~\cite{xu2022actformer,ghosh2022imos}, representing basic interactions such as "Shake Hands" or "Combat". 

\textbf{Spatial and Temporal Signal} Diverse spatial signals guide interactive motion generation, encompassing goal poses~\cite{taheri2024grip, lee2024interhandgen, zhang2024artigrasp, christen2022dgrasp, DIMOS}, root trajectory~\cite{braun2023physically, zhou2024gears, liu2023revisit, ugrinovic2024purposer}, root positions~\cite{kwon2024graspdiffusion, CIRCLE}, orientations~\cite{wang2024intercontrol}, object motions~\cite{li2023object}, and gamepad signals (e.g., instant direction, speed)~\cite{starke2019neural,starke2020local,starke2021neural}. These explicit, deterministic signals provide precise control over generated motions, enhancing both accuracy and adaptability while preserving motion fidelity.

\subsection{Fundamental Methods for Interaction Synthesis}
In this subsection, we introduce fundamental methods used in human interaction motion generation, ranging from classical approaches to the latest deep generative frameworks.

\subsubsection{Motion Graph}
Graph-based methods~\cite{kovar2023motion} represent a foundational approach in human interaction motion generation~\cite{Huang2014ActionReactionFT, Christos2018, Yang2020, Adeli_2021_tripod}, leveraging the inherent structure of motion data to synthesize novel sequences. These approaches typically implement motion graphs—directed graphs where nodes represent motion segments or poses, and edges denote viable transitions between segments. Novel motion synthesis occurs through graph traversal, where random walks along connected nodes generate coherent motion sequences. This framework enables the combinatorial fusion of characteristics from multiple exemplars, producing diverse motions while preserving the authenticity of the source data.
However, graph-based methods exhibit inherent limitations~\cite{kovar2023motion} in scalability. The approach necessitates storing the complete dataset and performing graph traversal during inference, introducing undesirable computational and storage overhead.

\subsubsection{Deterministic Regression}

Deterministic regression models~\cite{adeli2020socially, Kundu_2020_WACV, Baruah_2020_CVPRW} formulate interactive motion generation as a one-to-one mapping problem, predicting deterministic target motions from specified input conditions, typically supervised by L1 or L2 losses. These architectures commonly employ RNN~\cite{rumelhart1986rnn,cho2014GRU}, or Transformer~\cite{vaswani2023attentionneed} backbones to capture temporal dependencies via autoregressive regression. Nevertheless, their one-to-one mapping paradigm fails to capture the inherent stochasticity of human motions, often leading to mean poses and lifeless motions.

\subsubsection{Generative Adversarial Networks}
Generative Adversarial Networks (GANs)~\cite{goodfellow2020gan} have been commonly used for human interaction motion generation~\cite{men2021ganbasedreactivemg, xu2022actformer, goel2022interactionmixmatch}. The GAN architecture comprises two key components: a generator ($G$) and a discriminator ($D$). The generator ($G$) transforms random noise vectors sampled from a standard normal distribution ($\mathbf{z\sim\mathcal{N}(\mathbf{0}, \mathbf{I})}$) into interaction motions ($G(\mathbf{z})$). Meanwhile, the discriminator ($D$) evaluates the authenticity of the generated motions by learning to differentiate between real human motion samples ($\mathbf{x}$) from the training distribution and synthetic samples produced by $G$. This adversarial dynamic is formalized through the following objective function:
\begin{equation}
\begin{aligned}
\min_G \max_D \Bigl[
    &\, \mathbb{E}_{\mathbf{x} \sim p_{\text{data}}(\mathbf{x})}\bigl(\log D(\mathbf{x})\bigr) +\\
    &\, \mathbb{E}_{\mathbf{z} \sim p_{\mathbf{z}}(\mathbf{z})}\bigl(\log\bigl(1 - D(G(\mathbf{z}))\bigr)\bigr)
\Bigr],
\end{aligned}
\end{equation}

where \( p_{\text{data}}(\mathbf{x}) \) represents the distribution of real human motion data and \( p_{\mathbf{z}}(\mathbf{z}) \) denotes the prior distribution of the noise vector \( \mathbf{z} \). The generator seeks to minimize this objective by producing motions that the discriminator cannot reliably distinguish from real data, while the discriminator aims to maximize its ability to correctly classify real and generated motions.

Despite their impressive generative capabilities, GANs present training challenges~\cite{gui2021ganreview}. The inherent instability of adversarial training manifests itself in several critical issues: mode collapse, where the generator converges to produce only a limited subset of possible motions, and convergence problems, where the generator-discriminator dynamics fails to reach a stable equilibrium.

\subsubsection{Variational Autoencoders}
Variational Autoencoders (VAEs)~\cite{kingma2013vae} employ a two-stage architecture: first encoding input data into a structured latent space, then sampling from this learned distribution to reconstruct the original data. By maximizing the Evidence Lower Bound (ELBO), VAEs approximate the intractable log-likelihood of the data distribution. The ELBO for a VAE is expressed as:
\begin{equation}
L_{\theta, \phi}(\mathbf{x}) = \mathbb{E}_{\mathbf{z} \sim q_\phi(\mathbf{z}|\mathbf{x})} \left[ \ln p_\theta(\mathbf{x}|\mathbf{z}) \right] - D_{KL}\left(q_\phi(\mathbf{z}|\mathbf{x}) \parallel p_\theta(\mathbf{z})\right),
\end{equation}
where \( \mathbf{x} \) represents the input data, \( \mathbf{z} \) denotes the latent variables, \( \theta \) and \( \phi \) are the parameters of the decoder and encoder networks respectively, \( q_\phi(\mathbf{z}|\mathbf{x}) \) is the approximate posterior, and \( p(\mathbf{z}) \) is the prior distribution over the latent variables.

Conditional Variational Autoencoders (cVAEs)\cite{sohn2015cvae} extend the VAE framework by incorporating conditioning variables, enabling controlled data generation based on specific attributes. In human interaction motion synthesis, cVAEs have demonstrated versatility through various conditioning approaches: motion class labels\cite{Gupta_2023_DSAG, maheshwari2021mugl, ghosh2022imos, li2024task}, textual descriptions~\cite{milacski2024ghostgroundedhumanmotion, Narrator}, target poses~\cite{liu2023revisit, ugrinovic2024purposer}, and other control signals. The ELBO formulation is accordingly modified to incorporate the conditioning variable \( \mathbf{c} \):
\begin{equation}
\begin{aligned}
\mathcal{L}_{\theta, \phi}(\mathbf{x}|\mathbf{c}) = 
& \, \mathbb{E}_{\mathbf{z} \sim q_\phi(\mathbf{z}|\mathbf{x}, \mathbf{c})}
   \bigl[\ln p_\theta(\mathbf{x}|\mathbf{z}, \mathbf{c})\bigr] -\\
& \, D_{\mathrm{KL}}\!\Bigl(
    q_\phi(\mathbf{z}|\mathbf{x}, \mathbf{c})
    \,\Big\|\,
    p_\theta(\mathbf{z}|\mathbf{c})
\Bigr),
\end{aligned}
\end{equation}

where \( \mathbf{c} \) represents the conditioning information.

\subsubsection{Diffusion Models}
Diffusion models~\cite{ho2020denoising} have emerged as an expressive framework for generative modeling, demonstrating the capability to capture the complex data distribution in interactive human motions~\cite{shan2024opendomain, fan2024freemotion, ghosh2024remos, christen2024diffh2o, zhang2024hoidiffusion, cha2024text2hoi, LINGO, tesmo, AffordMotion}. The framework consists of two key processes: a forward diffusion process that systematically corrupts data with Gaussian noise across multiple timesteps until reaching a standard Gaussian distribution, and a learned reverse process that progressively denoises the corrupted data to reconstruct realistic human motions. The forward diffusion process is formally expressed as:
\begin{equation}
q(\mathbf{x}_t | \mathbf{x}_{t-1}) = \mathcal{N}(\mathbf{x}_t; \sqrt{1 - \beta_t} \mathbf{x}_{t-1}, \beta_t \mathbf{I}),
\end{equation}
where \( \beta_t \) represents the variance schedule at timestep \( t \), and \( \mathcal{N} \) denotes a Gaussian distribution. The reverse denoising process is modeled as:
\begin{equation}
p_\theta(\mathbf{x}_{t-1} | \mathbf{x}_t) = \mathcal{N}(\mathbf{x}_{t-1}; \mu_\theta(\mathbf{x}_t, t), \Sigma_\theta(\mathbf{x}_t, t)),
\end{equation}
where \( \mu_\theta \) and \( \Sigma_\theta \) are the mean and covariance predicted by the neural network parameterized by \( \theta \).

In contrast to GANs' single-step adversarial approach, the gradual, multi-step training dynamics of diffusion models provides inherent stability, enabling them to capture fine-grained motion details while maintaining diversity in their outputs.

\subsubsection{Transformer-Based Language Models}
Transformers~\cite{vaswani2023attentionneed} leverage self-attention mechanisms to efficiently capture long-range dependencies within data sequences. Introduced by Vaswani et al.~\cite{vaswani2023attentionneed}, the core self-attention equation is defined as:
\begin{equation}
\text{Attention}(Q, K, V) = \text{softmax}\left(\frac{QK^\top}{\sqrt{d_k}}\right)V,
\end{equation}
where \( Q \), \( K \), and \( V \) are the query, key, and value matrices derived from input embeddings, and \( d_k \) is the dimensionality of the key vectors. This mechanism allows the model to dynamically focus on relevant parts of the input sequence when generating each output element. 

Transformer-based language models, including GPTs~\cite{brown2020language, achiam2023gpt} and generative masked transformers~\cite{chang2022maskgit}, also become a promising paradigm for interaction motion modeling~\cite{siyao2024duolando,javed2024intermask}. These approaches typically implement a three-phase architecture: first, discretizing continuous motion data into tokens using encoders, like Vector Quantized Variational Autoencoders (VQ-VAEs)~\cite{van2017neuralvqvae}, which preserve essential motion structure and dynamics; second, modeling the sequential relationships between these tokens using transformer-based language models; and finally, projecting the tokenized representations back into continuous 3D motion sequences through a VQ-VAE decoder.

\subsubsection{RL + Physics Simulation}
Reinforcement Learning (RL) combined with physics simulation aims to generate more physically plausible interactive human motions~\cite{zhang2024artigrasp, christen2022dgrasp, unihsi, pan2023synthesizing, wang2024simssimulatinghumansceneinteractions, synthesizingphysicalcharactersceneinteractions}. This approach leverages RL's ability to learn optimal policies through trial-and-error while utilizing physics simulators to ensure that the generated motions adhere to fundamental physical laws. In this framework, an RL agent interacts with a physics-based environment, guided by rewards that promote target behaviors while accounting for constraints such as balance and collisions with objects or scenes. The design of reward functions plays a critical role in these approaches. On the one hand, RL-based methods often face training convergence challenges and exhibit limited generalization to novel actions. On the other hand, physics-based simulation remains essential as human motions inherently follow physical constraints in the real world—a fundamental aspect that purely kinematic-based methods struggle to capture.

\begin{figure*}
	\begin{center}
		\includegraphics[width=0.95\linewidth]{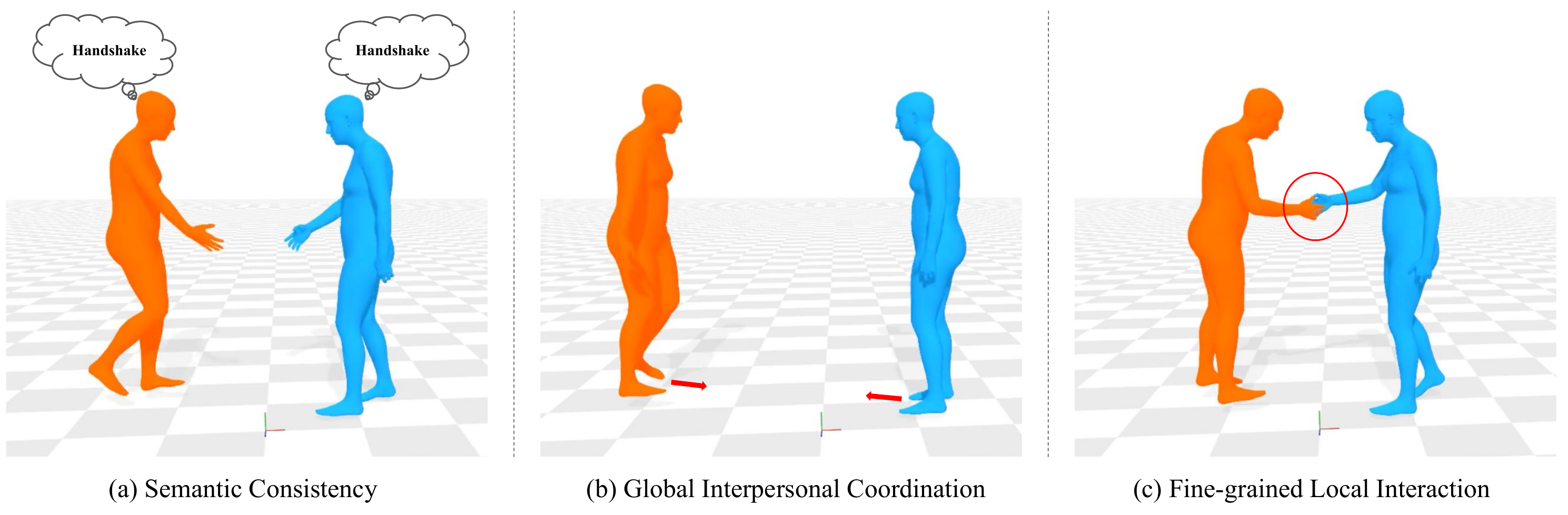}
	\end{center}
	\vspace{-10pt}
	\caption{Illustration of three major challenges in human-human interaction generation: (a) Semantic consistency; (b) Global interpersonal coordination; and (c) Fine-grained local interaction. All figures are adapted from~\cite{xu2023interx}.}
	\label{fig:hh_challenges_viz}
	\vspace{-10pt}
\end{figure*}

\subsubsection{LLM-Based Motion Planning}
Recent advances in Large Language Models (LLMs)\cite{minaee2024llmsurvey} have enabled their application as automated motion planners\cite{chopin2023interformer, unihsi, xu2024interdreamer}, translating high-level interaction goals into detailed step-by-step motion sequences. These approaches innovate by generating interactions without relying on extensive interaction datasets, instead leveraging the knowledge of human kinematics embedded in pre-trained LLMs through carefully designed prompts. LLMs can provide temporal specifications of interactions, identify relevant joint involvement, and describe precise interaction dynamics.

\section{Human Interaction Motion Generation}
\label{sec:interactive-hmg}

This section elaborates on solutions for generating interactive human motions across four categories, as illustrated in Fig.~\ref{fig:interaction_viz}: human-human interactions (Sec.~\ref{sec:interactive-hh-mg}), human-object interactions (Sec.~\ref{sec:interactive-ho-mg}), human-scene interactions (Sec.~\ref{sec:interactive-hs-mg}), and multi-entity interactions involving combinations of these elements (Sec.~\ref{sec:interactive-hm-mg}).

\vspace{-6pt}
\subsection{Human-Human Interaction Generation}
\label{sec:interactive-hh-mg}

Human-human interaction (HHI) research spans from dyadic interactions to group dynamics. For two-person interactions, the literature addresses two primary tasks: actor-reactor generation~\cite{ghosh2024remos,siyao2024duolando}, which synthesizes reactive motions in response to an actor's movements, and synchronized interaction generation~\cite{javed2024intermask,InterGen}, which simultaneously creates coordinated motions for both participants. Research also extends beyond dyadic interactions to group scenarios~\cite{fan2024freemotion,shan2024opendomain}, addressing the synthesis of coordinated movements among multiple participants.

Generating realistic HHI motions requires addressing three critical aspects, as illustrated in Fig.~\ref{fig:hh_challenges_viz}: semantic consistency, global interpersonal coordination, and fine-grained local interaction.
Human interactions are inherently intention-driven. \textit{Semantic consistency} ensures coherent and meaningful movements between participants that align with these intentions. For example, in social interactions like Fig.~\ref{fig:hh_challenges_viz}~(a), body movements serve specific social gestures (e.g., handshakes), while in the duet dance, movements synchronize with musical cues. To enforce semantic consistency, existing approaches incorporate conditioning signals, such as textual descriptions or music.
\textit{Global interpersonal coordination} manages natural spatial relationships between interacting individuals (Fig.~\ref{fig:hh_challenges_viz}~(b)), including relative positions, mutual orientations, and interpersonal distances across scenarios. This coordination requires each person's global position to be responsive to their partner's movements, enabling fluid and harmonious interactions.
\textit{Fine-grained local interaction} addresses precise close-proximity interactions between body parts, such as hand-holding (Fig.~\ref{fig:hh_challenges_viz}~(c)) and hugging. This aspect focuses on encouraging realistic contact while minimizing artifacts like interpenetration between bodies. A comprehensive analysis of HHI generation methods is provided in Table~\ref{table:hhmg_methods}.

\subsubsection{Modeling Semantics of HHI}
\label{sec:interactive-hh-mg-sc}

\begin{table*}[t]
	\centering
	\captionsetup{font=small}
	\caption{Representative works of human-human interaction motion generation.}
	\label{table:hhmg_methods}
	\vspace{-5pt}
	\begin{threeparttable}
	\resizebox{0.99\textwidth}{!}{
	\setlength\tabcolsep{2.5pt}
	\renewcommand\arraystretch{1.02}
	\begin{tabular}{r||c|c|c|c|c|c}
	\hline\thickhline
	\rowcolor{mygray1}
        Method & Year & Venue & Model & Condition & Multi-Human Interaction Dataset & Single-Human Dataset\\
	\hline
	\hline
        Shan et al.~\cite{shan2024opendomain} & 2024 & ECCV & Diffusion & Text Description & LAION-Pose~\cite{shan2024opendomain}, WebVid-Motion~\cite{shan2024opendomain}, InterHuman~\cite{InterGen} & HumanML3D~\cite{HumanML3D} \\
        
        FreeMotion~\cite{fan2024freemotion} & 2024 & ECCV & Diffusion & Actor's Motion, Spatial Signal, Text & InterHuman~\cite{InterGen} & - \\
        
        ReMoS~\cite{ghosh2024remos} & 2024 & ECCV & Diffusion & Actor's Motion & ReMoCap~\cite{ghosh2024remos}, ExPI~\cite{guo2022multipersonexmotion}, 2C~\cite{2C}, InterHuman~\cite{InterGen} & - \\
        
        ReGenNet~\cite{xu2024regennet} & 2024 & CVPR & Diffusion & Actor's Motion & NTU RGB+D 120~\cite{Liu_2020_NTURGBD120}, InterHuman~\cite{InterGen}, Chi3D~\cite{Fieraru_2020_CVPR} & - \\
        
        In2IN~\cite{ponce2024in2in} & 2024 & CVPRW & Diffusion & Text Description & InterHuman~\cite{InterGen} & HumanML3D~\cite{HumanML3D} \\
        
        InterControl~\cite{wang2024intercontrol} & 2024 & NeurIPS & Diffusion & Spatial Signal, Text Description & - & HumanML3D~\cite{HumanML3D}, KIT-ML~\cite{KIT-ML} \\
        
        InterGen~\cite{InterGen} & 2024 & IJCV & Diffusion & Text Description & InterHuman~\cite{InterGen} & - \\
        
        PriorMDM~\cite{shafir2023ComMDM} & 2024 & ICLR & Diffusion & Text Description & 3DPW~\cite{Marcard_2018_ECCV} & HumanML3D~\cite{HumanML3D}, BABEL~\cite{BABEL}\\
        
        Duolando~\cite{siyao2024duolando} & 2024 & ICLR & LLM & Actor's mMtion, Audio & DD100~\cite{siyao2024duolando} & - \\
        
        Social Diffusion~\cite{SocialDiffusion} & 2023 & ICCV & Diffusion & Past Motion & Haggling~\cite{joo2019cvpr}, MuPoTS-3D~\cite{mehta2018ssmp}, 3DPW~\cite{Marcard_2018_ECCV} & - \\
        
        ActFormer~\cite{xu2022actformer} & 2023 & ICCV & GAN, Transformer & Action Class & NTU RGB+D 120~\cite{Liu_2020_NTURGBD120}, GTA Combat~\cite{xu2022actformer} & BABEL~\cite{BABEL} \\
        
        InterFormer~\cite{chopin2023interformer} & 2023 & TMM & Transformer & Actor's Motion & SBU Kinect~\cite{sbu_kinect}, K3HI~\cite{K3HI}, DuetDance~\cite{Kundu_2020_WACV} & - \\
        
        Tanaka and Fujiwara~\cite{Tanaka_2023_ICCV} & 2023 & ICCV & Transformer & Text Description & NTU RGB+D 120~\cite{Liu_2020_NTURGBD120} & - \\
        
        DSAG~\cite{Gupta_2023_DSAG} & 2023 & WACV & cVAE & Action Class, Past Motion & NTU RGB+D 120~\cite{Liu_2020_NTURGBD120} & HumanAct12~\cite{guo2020action2motion}, UESTC~\cite{UESTC}, Human3.6M~\cite{ionescu2013human3} \\
        
        SoMoFormer~\cite{vendrow2022somoformer} & 2023 & WACV & Transformer & Past Motion & 3DPW~\cite{Marcard_2018_ECCV}, CMU-MoCap~\cite{cmu_mocap}, MuPoTS-3D~\cite{mehta2018ssmp} & AMASS~\cite{AMASS} \\
        
        Guo et al.~\cite{guo2022multipersonexmotion} & 2022 & CVPR & Transformer & Past Motion & ExPI~\cite{guo2022multipersonexmotion} & - \\
        
        MUGL~\cite{maheshwari2021mugl} & 2022 & WACV & cVAE & Action Class, Past Motion & NTU RGB+D 120~\cite{Liu_2020_NTURGBD120} & - \\
        
        Interaction Mix and Match~\cite{goel2022interactionmixmatch} & 2022 & SCA & GAN & Action Class, Actor's Motion & SBU Kinect~\cite{sbu_kinect}, 2C~\cite{2C} & - \\
        
        Men et al.~\cite{men2021ganbasedreactivemg} & 2022 & C\&G & GAN & Actor's Motion & SBU Kinect~\cite{sbu_kinect}, HHOI~\cite{shu2016HHOI}, 2C~\cite{2C} & - \\
        
        TRiPOD~\cite{Adeli_2021_tripod} & 2021 & ICCV & Graph, Regression & Past Motion, Video & 3DPW~\cite{Marcard_2018_ECCV} & - \\
        
        MRT~\cite{wang2021MRT} & 2021 & NeurIPS & Transformer & Past Motion & 3DPW~\cite{Marcard_2018_ECCV}, CMU-MoCap~\cite{cmu_mocap}, MuPoTS-3D~\cite{mehta2018ssmp} & - \\
        
        Baruah and Banerjee~\cite{Baruah_2020_CVPRW}\ & 2020 & CVPRW & Agent (Regression) & Past Motion & SBU Kinect~\cite{sbu_kinect}, K3HI~\cite{K3HI} & - \\
        
        Yang et al.~\cite{Yang2020} & 2020 & SIGGRAPH & Graph & Audio & Private Dataset & - \\
        
        Kundu et al.~\cite{Kundu_2020_WACV} & 2020 & WACV & Regression & Past Motion & DuetDance~\cite{Kundu_2020_WACV}, CMU-MoCap~\cite{cmu_mocap}, SBU Kinect~\cite{sbu_kinect} & - \\
        
        Adeli et al.~\cite{adeli2020socially} & 2020 & RA-L & Regression & Past Motion, Video & NTU RGB+D 120~\cite{Liu_2020_NTURGBD120} & - \\
        
        Joo et al.~\cite{joo2019cvpr} & 2019 & CVPR & AE & Actor's Motion & Haggling~\cite{joo2019cvpr} & - \\
        
        Ahuja et al.~\cite{ahuja2019reactornot} & 2019 & ICMI & Attention-Based & Actor's Motion, Audio & Private dataset & - \\
        
        Christos Mousas~\cite{Christos2018} & 2018 & VR & Graph & Actor's Motion & CMU-MoCap~\cite{cmu_mocap} & - \\
        
        Action-Reaction~\cite{Huang2014ActionReactionFT} & 2014 & ECCV & Graph & Video & UT-Interaction~\cite{UT-Interaction-Data}, SBU Kinect~\cite{sbu_kinect} & - \\
	\hline
	\end{tabular}
	}
	\end{threeparttable}
	\vspace{-5pt}
\end{table*}

Human interactions encompass diverse semantic contexts across varying scenarios. This section reviews existing approaches for interaction generation based on their contextual constraints, including past interactions, partner motions, action categories, texts, and audio.

\paragraph{Conditioning on Past Interactions.} Early research on HHI generation focused on predicting future interactive motions using past interaction motions. Kundu et al.~\cite{Kundu_2020_WACV} propose a motion generation architecture that alternates between two fixed-length cross-person motion prediction recurrent models to generate actor and reactor motions sequentially. This approach enables long-term motion synthesis while maintaining temporal synchronization within each window. Baruah and Banerjee~\cite{Baruah_2020_CVPRW} develop a predictive agent with two core components: an observation module that dynamically identifies salient body parts from previous motions and encodes them as pose features, and a recurrent motion completion module that predicts subsequent poses. This architecture facilitates the generation of contextually appropriate interactive motions. MRT~\cite{wang2021MRT} introduces a multi-range transformer comprising local-range and global-range encoders. The local-range encoder extracts individual motion features, while the global-range encoder captures inter-person interaction features. This architecture ensures semantic consistency and scales to both dyadic and multi-person scenarios, supporting up to 15 individuals by decoupling local features from global interactions. Similarly, Guo et al.~\cite{guo2022multipersonexmotion} and SoMoFormer~\cite{vendrow2022somoformer} leverage transformer architectures for semantic consistency in motion generation. Guo et al. implement a cross-interaction attention module for bidirectional motion information exchange, while SoMoFormer employs learned identity embeddings to maintain semantic alignment between participants. Social Diffusion~\cite{SocialDiffusion} adopts a recurrent diffusion model for simultaneous multi-person motion generation, incorporating an order-invariant averaging function to aggregate motion features while preserving social roles and interaction dynamics.

\paragraph{Conditioning on Partner Motions.} Partner motion conditioning, also known as reactive motion synthesis, focuses on generating one person's motion in response to their partner's motion state. Christos Mousas~\cite{Christos2018} proposes a real-time HMM-based model for virtual dance partner generation. The model represents dance sequences as a graph of paired actor-reactor motion vectors, maintaining semantic alignment between roles. The HMM supports jump-state transitions to accommodate improvisation, enabling the virtual reactor to adapt dynamically to the actor's movements.
Sebastian et al.~\cite{starke2021neural,starke2020local} predict subsequent poses using a gated mixture of expert networks that processes phase features extracted from both the character and its opponent's current poses. Men et al.~\cite{men2021ganbasedreactivemg} develop a GAN-based approach, incorporating a temporal seq2seq attention module in the generator to preserve semantic consistency across key temporal steps. Building on this, InterFormer~\cite{chopin2023interformer} implements a Transformer with temporal and spatial attention mechanisms, introducing a first-frame loss function to align initial poses. This alignment is particularly valuable given the limited availability of long-distance interaction training data.
ReGenNet~\cite{xu2024regennet} adopts a diffusion model for iterative reactor motion generation, employing a simple yet effective strategy of concatenating actor motion features with generated reactor motion at each timestamp. ReMoS~\cite{ghosh2024remos} extends this diffusion-based framework by incorporating spatio-temporal cross-attention and hand-interaction-aware cross-attention. These additions enable the model to analyze correlations between actor and reactor motion features, resulting in synchronized whole-body interactions with enhanced semantic consistency.

\paragraph{Conditioning on Action Categories.} Action class labels provide a approach to encode high-level motion semantics. MUGL~\cite{maheshwari2021mugl} and DSAG~\cite{Gupta_2023_DSAG} implement deep learning encoder-decoder architectures based on conditional Gaussian Mixture VAE~\cite{dilokthanakul2016deep}, where encoded features are modulated by action class labels to capture semantic relationships across participants' motions.
ActFormer~\cite{xu2022actformer} employs a GAN-based Transformer framework that converts latent vector sequences and action labels into multi-person motion sequences. The model represents action classes as token embeddings combined with latent vectors, ensuring semantic consistency across generated motions. Interaction Mix and Match~\cite{goel2022interactionmixmatch} advances reactive motion synthesis using a GAN with multi-hot class embeddings, enabling the generation of reactor motions using single-class or designed class combinations.

\paragraph{Conditioning on Texts.} Recent advances enable text-driven human interaction motion generation. Tanaka et al.~\cite{Tanaka_2023_ICCV} approach this task using a transformer model that encodes paired text prompts in active and passive voices for actor and reactor roles, effectively maintaining distinct semantic roles throughout asymmetric interactions. Several subsequent works~\cite{shafir2023ComMDM, InterGen, wang2024intercontrol, ponce2024in2in} adopt diffusion-based approaches with text conditioning.

PriorMDM~\cite{shafir2023ComMDM} introduces a lightweight communication block to coordinate two frozen Motion Diffusion Models (MDMs)~\cite{tevet2023MDM}, widely used in text-to-motion generation, enabling two-person motion generation while preserving pre-trained model's capabilities. Building on this, InterGen~\cite{InterGen} incorporates CLIP-encoded text prompts and a mutual attention mechanism in two transformer-based denoisers with shared weights. This shared contextual understanding prevents mode collapse, where generated interactions may result in unrelated motion semantics between participants.
In2IN~\cite{ponce2024in2in} implements multi-head cross-attention modules within a transformer-based Siamese diffusion model for refined interaction control. The model employs a multi-weight CFG strategy that independently weights interaction and individual description influences, enabling precise control over movement diversity while maintaining semantic coherence. InterMask~\cite{javed2024intermask} takes a different approach, encoding two-person motions into 2D discrete token maps and using generative masked modeling for text-to-motion generation.
Recent work has expanded beyond dyadic interactions to handle multi-person scenarios~\cite{shan2024opendomain,fan2024freemotion}. Notably, Shan et al.~\cite{shan2024opendomain} propose a unified transformer-based diffusion framework featuring interleaved pose and motion layers. Pose layers, conditioned on text embeddings, ensure semantic consistency, while motion layers, conditioned on individual poses, maintain temporal alignment between participants.

\paragraph{Conditioning on Audio.} Interactive human motion can also be generated from audio signals, such as conversational speech and music. Unlike text conditioning, audio-driven generation requires precise temporal alignment. Ahuja et al.~\cite{ahuja2019reactornot} introduce a dyadic residual attention model that combines two components: a monadic module processing only the reactor's audio and motion, and a dyadic module integrating both participants' audio, motion, and predicted reactor dynamics. This architecture generates responsive reactor movements that align with both actor motion and audio cues.
Yang et al.\cite{Yang2020} develop a motion graph incorporating audio-motion coordination features including phonemic clause alignment, speaker hesitations, and listener responses. This approach enables the generation of synchronized multi-human motions that coherently follow conversational dynamics. Duolando\cite{siyao2024duolando} presents a GPT-based generative model to predict follower dance motions autoregressively, guided by both music and leader motion. The model processes audio features extracted via Librosa~\cite{mcfee2015librosa}, mapping them to match motion feature dimensions. This design ensures balanced semantic alignment between musical and motion elements.

\subsubsection{Maintaining Global Interpersonal Coordination}
\label{sec:interactive-hh-mg-gic}
Maintaining accurate global spatial alignment and mutual orientation between interacting individuals is crucial for realistic interaction generation. Recent models have addressed this challenge through various approaches.

Several works focus on spatial guidance and loss functions. ReMoS~\cite{ghosh2024remos} implements a spatial guidance function during inference to align reactor movements with actor positions, enhancing spatial coherence. ReGenNet~\cite{xu2024regennet} introduces a distance-based interaction loss that minimizes discrepancies in body poses, orientations, and translations. InterControl~\cite{wang2024intercontrol} extends MDM~\cite{tevet2023MDM} by incorporating Motion ControlNet~\cite{zhang2023controlnet} to condition the model on spatial control signals, improving temporal alignment.

Other approaches emphasize representation and encoding strategies. InterGen~\cite{InterGen} employs a non-canonical motion representation that encodes global joint positions and orientations in a unified world frame. SoMoFormer~\cite{vendrow2022somoformer} integrates grid positioning embeddings to encode global locations, effectively capturing spatial relationships between participants. MUGL~\cite{maheshwari2021mugl} and DSAG~\cite{Gupta_2023_DSAG} implement global trajectory encoder-decoder modules with specialized loss functions to enforce movement path alignment.

Some models incorporate additional learning techniques. Duolando~\cite{siyao2024duolando} utilizes off-policy reinforcement learning to improve the handling of novel leader motions, reducing unrealistic displacements and skating artifacts. Joo et al.~\cite{joo2019cvpr} propose an autoencoder-based network for haggling scenarios that incorporates both body poses and face orientations to improve spatial alignment with ground truth.

\begin{figure}[t]
	\centering
	\includegraphics[width=0.48\textwidth]{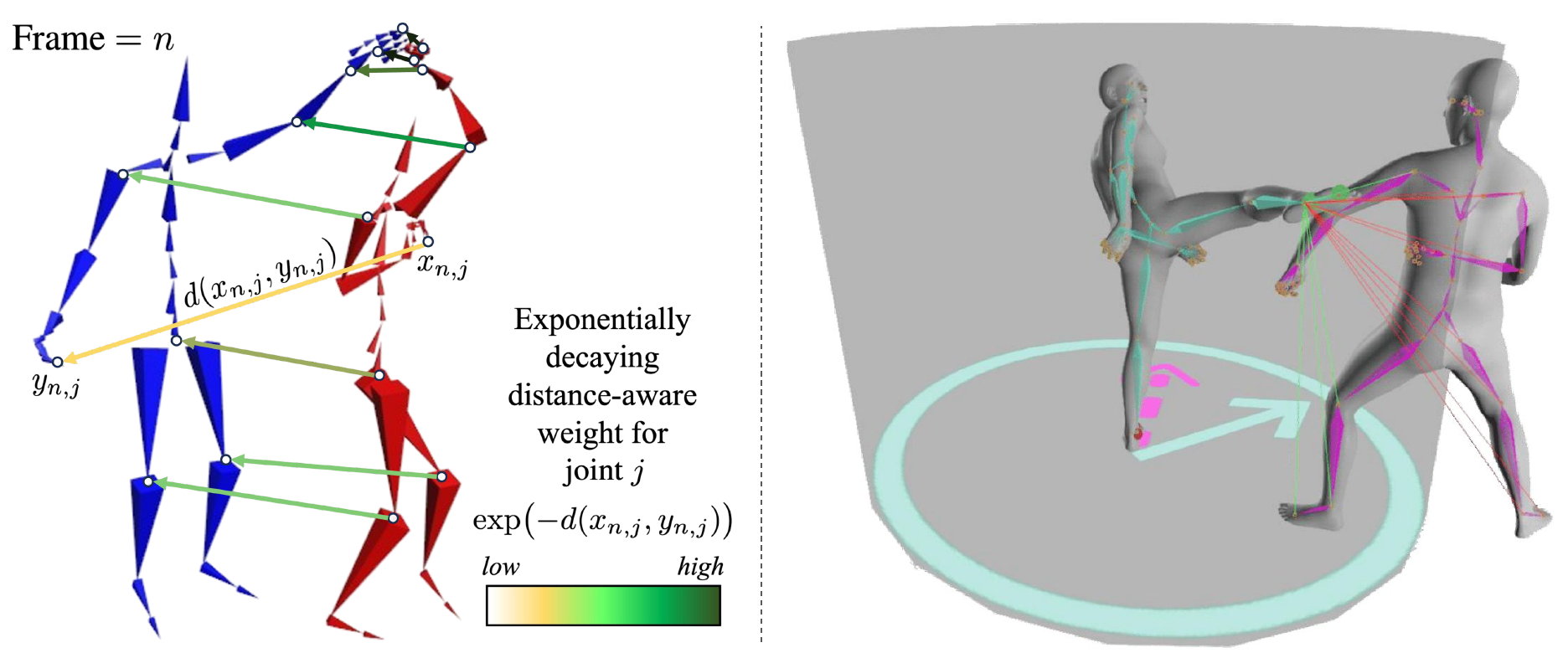}
	\vspace{-10pt}
	\caption{\small Examples of distance-aware interaction mechanisms. (Left) ReMoS’s~\cite{ghosh2024remos} Distance-Aware Reaction Loss applies an exponentially decaying function to prioritize reactor joints closer to the actor. (Right) InterGen~\cite{InterGen} introduces Joint Distance Loss, which activates only when the horizontal distance between two individuals falls within a specified range, represented by the cylindrical region in the figure. Figures are adapted from~\cite{ghosh2024remos,InterGen}.}
	\label{fig:hh_distance_loss}
	\vspace{-10pt}
\end{figure}

\subsubsection{Capturing Fine-Grained Local Interaction}
\label{sec:interactive-hh-mg-fli}
Fine-grained local interactions, such as handshakes and face touching, require precise modeling of synchronized and complementary movements between proximate body parts of interacting individuals. Recent models have developed various strategies to enhance physical contact accuracy and interaction realism.

Several approaches focus on distance-based mechanisms. ReMoS~\cite{ghosh2024remos} implements an exponentially decaying distance-aware reaction loss that prioritizes reactor joints closer to the actor, improving the fidelity of close-range interaction, as illustrated in Fig.~\ref{fig:hh_distance_loss}~(Left). InterGen~\cite{InterGen} employs a joint distance map loss that activates only within specific proximity thresholds, strengthening physical connections during interactions, as illustrated in Fig.~\ref{fig:hh_distance_loss}~(Right). InterFormer~\cite{chopin2023interformer} incorporates an interaction distance module that integrates softmax-scaled joint distances into the attention matrix, effectively modeling fine-grained interactions by weighting closer joints more heavily.

Architectural and strategy designs are addressed in other works. Duolando~\cite{siyao2024duolando} utilizes a VQ-VAE to model relative joint translations between participants, enhancing the authenticity of physical contact. InterControl~\cite{wang2024intercontrol} guides motion generation using LLM-generated interaction descriptions that specify contact points and temporal progression. 

\vspace{-6pt}
\subsection{Human-Object Interaction Generation}
\label{sec:interactive-ho-mg}
Human-object interaction (HOI) generation aims to synthesize realistic, context-aware motion sequences depicting humans engaging with 3D objects. This capability enhances system functionality and user immersion in robotics, virtual reality, human-computer interaction, and computer graphics applications.

Current research primarily addresses two key challenges: interaction semantic relevance, and spatial and physical constraints.
\textit{Interaction semantic relevance} ensures that generated motions authentically reflect interaction intents and proper object functionality—for example, `lifting' a book or `sitting' on a chair—while maintaining plausibility and fidelity. \textit{Spatial and physical constraints} take charge of adherence to physical laws and contact dynamics, requiring precise alignment between human body parts and object geometries while respecting gravity, friction, and biomechanical limits. 

These challenges are being addressed through sophisticated data-driven models, advanced physics-based simulations, and multimodal data integration, establishing 3D HOI generation as a rapidly evolving field in computer vision and graphics. A comprehensive analysis of human-object interaction generation methods is presented in Table~\ref{table:homg_methods}. This section reviews recent innovations in semantic relevance and constraint modeling.

\begin{table*}[t]
	\centering
	\captionsetup{font=small}
	\caption{Representative works of human-object interaction motion generation.}
	\label{table:homg_methods}
	\vspace{-5pt}
	\begin{threeparttable}
	\resizebox{0.99\textwidth}{!}{
	\setlength\tabcolsep{2.5pt}
	\renewcommand\arraystretch{1.02}
	\begin{tabular}{r||c|c|c|c|c}
	\hline\thickhline
	\rowcolor{mygray1}
        Method & Year & Venue & Model & Condition & Human-Object Interaction Dataset \\
	\hline
	\hline
        DiffH2O\cite{christen2024diffh2o} & 2024 & SIGGRAPH Asia & Diffusion & Text Description & GRAB\cite{taheri2020grab} \\

        InterDreamer\cite{xu2024interdreamer} & 2024 & NeurIPS & LLM & Text Description & BEHAVE~\cite{bhatnagar2022behave}, CHAIRS~\cite{jiang2023full} \\
        
        CHOIS\cite{li2025controllable} & 2024 & ECCV & Diffusion & Text Description, Sparse Object Waypoints & OMOMO\cite{li2023object} \\
        
        HIMO-Gen\cite{lv2024himo} & 2024 & ECCV & Diffusion & Text Description & HIMO\cite{lv2024himo} \\
        
        InterFusion\cite{dai2024interfusion} & 2024 & ECCV & GPT-4V, Optimization & Text Description & ChatGPT prompts, DeepFloyd  \\
        
        F-HOI\cite{yang2024f} & 2024 & ECCV & Multimodal LLM & Text Description & Semantic-HOI~\cite{yang2024f} \\
        
        GraspDiff\cite{zuo2024graspdiff} & 2024 & TVCG & Diffusion & Object Shape & ObMan\cite{hasson19_obman}, HO3D\cite{hampali2020honnotate}, FPHAB\cite{garcia2018first} \\

        HOIDiffusion\cite{zhang2024hoidiffusion} & 2024 & CVPR & Diffusion & Object Model, Text Description& DexYCB\cite{chao2021dexycb} \\
        
        Text2HOI\cite{cha2024text2hoi} & 2024 & CVPR & Diffusion & Text Description & GRAB\cite{taheri2020grab}, ARCTIC\cite{fan2023arctic}, H2O\cite{Kwon_2021_ICCV} \\
        
        CG-HOI\cite{diller2023cghoi} & 2024& CVPR & Diffusion & Text Description& BEHAVE\cite{bhatnagar2022behave}, CHAIRS\cite{jiang2023full} \\

        GEARS\cite{zhou2024gears} & 2024 & CVPR & MLP, Attention & Hand-Object Trajectory, Object Shape & GRAB\cite{taheri2020grab}, InterCap\cite{huang2022intercap}, ObMan\cite{hasson19_obman}\\
        
        G-HOP\cite{ye2023ghop} & 2024 & CVPR & Diffusion & Hand-Object Interaction Images, Object Mesh & ContactPose\cite{brahmbhatt2020contactpose}, DexYCB\cite{chao2021dexycb}, YCB-Affordance\cite{corona2020ganhand}, HOI4D\cite{Liu_2022_CVPR}, GRAB\cite{taheri2020grab}, OakInk\cite{yang2022oakink}  \\

        InterHandGen~\cite{lee2024interhandgen} & 2024 & CVPR & Diffusion & Hand Pose, Object Shape (Optional) & ARCTIC\cite{fan2023arctic} , InterHand2.6M\cite{moon2020interhand2}\\

        NIFTY\cite{kulkarni2024nifty} & 2024 & CVPR & Diffusion & Initial Body Pose, Object Shape & BEHAVE\cite{bhatnagar2022behave}  \\

        IM-HOI\cite{zhao2024imhoi} & 2024 & CVPR & Diffusion & IMU Recording, Monocular RGB Video & IMHD2\cite{zhao2024imhoi}, BEHAVE\cite{bhatnagar2022behave}, InterCap\cite{huang2022intercap}, CHAIRS\cite{jiang2023full}, HODome\cite{zhang2023neuraldome}  \\

        GeneOHDiffusion\cite{liu2024geneoh} & 2024 & ICLR & Diffusion & Noisy Hand-Object Poses & GRAB\cite{taheri2020grab}, HOI4D\cite{Liu_2022_CVPR}, ARCTIC\cite{fan2023arctic}  \\
        
        ArtiGrasp\cite{zhang2024artigrasp} & 2024 & 3DV & RL, Physics Simulation & Hand Pose Reference & ARCTIC\cite{fan2023arctic} \\
        
        GRIP\cite{taheri2024grip} & 2024 & 3DV & RNN & Body Pose, Object Motion & GRAB\cite{taheri2020grab}, InterCap\cite{huang2022intercap}, MoGaze\cite{kratzer2020mogaze}\\

        MACS\cite{MACS2024} & 2024 & 3DV & Diffusion & Action Label, Object Mass & ManipNet Data\cite{zhang2021manipnet} \\

        PhysFullbody\cite{braun2023physically} & 2024 & 3DV & RL & Hand-Object Reference Pose, Wrist Trajectory & GRAB\cite{taheri2020grab} \\
      
        TOHO\cite{li2024task} & 2024 & WACV & cVAE & Task Instructions & GRAB\cite{taheri2020grab} \\
        
        CWGrasp\cite{paschalidis20243d} & 2024 & - & cVAE, Optimization & Object Mesh & ReplicaGrasp\cite{tendulkar2023flex}, GRAB\cite{taheri2020grab}, CIRCLE\cite{CIRCLE}\\

        GraspDiffusion\cite{kwon2024graspdiffusion} & 2024 & - & Diffusion & Object Mesh and Position & GRAB\cite{taheri2020grab}, BEHAVE\cite{bhatnagar2022behave}, DexYCB\cite{chao2021dexycb}, HICO-DET\cite{chao2018learning}, V-COCO\cite{gupta2015visual}\\

        HOI-Diff\cite{peng2023hoi} & 2023 & - & Diffusion & Text Description & BEHAVE\cite{bhatnagar2022behave}, OMOMO\cite{li2023object} \\

        OMOMO\cite{li2023object} & 2023 & SIGGRAPH Asia & Diffusion & Object Motions & OMOMO\cite{li2023object}  \\
                
        InterDiff\cite{xu2023interdiff} & 2023 & ICCV & Diffusion & Initial Pose & BEHAVE\cite{bhatnagar2022behave}  \\

        Chen et al.~\cite{chen2023synthesizing} & 2023 & SIGGRAPH & cVAE, Optimization & Object Shape & YCB\cite{calli2015ycb}\\
        
        CAMS\cite{Zheng_2023_CVPR} & 2023 & CVPR & cVAE & Init. Hand Pose, Object Goal Sequence, Object Shape & HOI4D\cite{Liu_2022_CVPR}  \\

        FLEX\cite{tendulkar2023flex} & 2023 & CVPR & MLP & Object Shape & GRAB\cite{taheri2020grab}, ReplicaGrasp\cite{tendulkar2023flex}  \\
        
        NeuralDome\cite{zhang2023neuraldome} & 2023 & CVPR & Neural Rendering & Multi-View Video Sequence & HODome\cite{zhang2023neuraldome}  \\

        SAGA\cite{wu2022saga} & 2022 & ECCV & cVAE & Human-Object Initial Pose, Object Shape & GRAB\cite{taheri2020grab}, HO3D\cite{hampali2020honnotate}, AMASS\cite{AMASS}\\
        
        TOCH\cite{zhou2022toch} & 2022 & ECCV & GRU & Hand-Object Image / Hand-Object Mesh & GRAB\cite{taheri2020grab}, HO3D\cite{hampali2020honnotate} \\
        
        DGrasp~\cite{christen2022dgrasp} & 2022 & CVPR & RL, Physics Simulation & Grasp Label (Image), Object Pose & DexYCB\cite{chao2021dexycb}, HO3D\cite{hampali2020honnotate}  \\
        
        GOAL\cite{taheri2021goal} & 2022 & CVPR & cVAE, Optimization & Human-Object Initial Pose, Object Shape & GRAB\cite{taheri2020grab}  \\
        
        IMoS\cite{ghosh2022imos} & 2022 & Eurographics & cVAE, Optimization & Action Labels, Object & GRAB\cite{taheri2020grab} \\

        GraviCap\cite{GraviCap2021} & 2021 & ICCV & Regression, Optimization & Monocular RGB Videos & Gravicap Dataset \\
        
        ManipNet~\cite{zhang2021manipnet} & 2021 & SIGGRAPH & RNN & Object Shape, Wrist-Object Trajectories & Object Manipulation Dataset  \\
                
        GrabNet~\cite{taheri2020grab} & 2020 & ECCV & cVAE, RNN & Object Shape & GRAB~\cite{taheri2020grab}  \\

        GanHand\cite{corona2020ganhand} & 2020 & CVPR & RNN & Single RGB Image & ObMan\cite{hasson19_obman}, YCB-Affordance \cite{corona2020ganhand}  \\

	\hline
	\end{tabular}
	}
	\end{threeparttable}
	\vspace{-5pt}
\end{table*}

\subsubsection{Modeling Semantic of Human-Object Interaction}
\label{sec:interactive-ho-mg-if}

Early approaches like GOAL~\cite{taheri2021goal} and SAGA~\cite{wu2022saga} focus on generating whole-body motions for object grasping, while IMoS~\cite{ghosh2022imos} extends this by incorporating action categories to guide the grasping process. These approaches typically adopt a two-stage pipeline, wherein full-body motion is first synthesized using auto-regressive models or conditional variational autoencoders, followed by grasp motion optimization.

Recent works have leveraged diffusion models to generate human interactions with objects. Nifty~\cite{kulkarni2024nifty} and COUCH~\cite{zhang2022couch} demonstrate this approach for human-furniture interactions. Many subsequent works utilize textual descriptions as conditioning signals. DiffH2O~\cite{christen2024diffh2o} introduces a two-stage diffusion process that decouples grasping and interaction phases while ensuring smooth transitions through grasp guidance. It employs a canonicalized hand-object representation incorporating distances between hand joints and their nearest points on the object mesh. Similarly, HOI-Diff~\cite{peng2023hoi} generates coarse-level human and object motions before estimating contact points through an affordance prediction diffusion model. GRIP~\cite{taheri2024grip} employs separate networks for arm denoising and body-hand motion consistency. HOIDiffusion~\cite{zhang2024hoidiffusion} refines a controllable Stable Diffusion model using geometric structures from grasp trajectories, enabling precise pose control. Text2HOI~\cite{cha2024text2hoi} introduces a contact map prediction stage from canonical object meshes and text prompts, followed by denoising and refinement for accurate single-handed and dual-handed interactions. CHOIS~\cite{li2025controllable} utilizes object geometry through Basis Point Set (BPS) representation~\cite{prokudin2019efficient} alongside language descriptions. HIMO-Gen~\cite{lv2024himo} employs separate text-conditioned diffusion models for human and object motion, connected through a mutual interaction module. InterFusion~\cite{dai2024interfusion} learns image-pose mappings to understand and generate various HOI scenes, while F-HOI~\cite{yang2024f} leverages multimodal large language models for fine-grained HOI generation tasks. InterDreamer\cite{xu2024interdreamer} achieves zero-shot text-conditioned generation by using LLMs for high-level task planning.

A distinct category of methods focuses on object-guided interaction synthesis. OMOMO~\cite{li2023object} generates full-body manipulation behaviors solely from object geometries and motions. TOHO~\cite{li2024task} estimates object end positions from task descriptions which then guides the generation of human pose sequences. GraspDiffusion~\cite{kwon2024graspdiffusion} employs a two-stage pipeline for synthesizing whole-body interactions given object mesh and its position information.

\subsubsection{Spatial and Physical Constraints}
Physical plausibility through accurate contact modeling is fundamental in HOI motion synthesis. Recent approaches have addressed this challenge through various contact-aware methodologies. This subsection also extends the discussion to encompass HOI pose generation for a comprehensive review.

\begin{figure}[t]
	\centering
	\includegraphics[width=0.48\textwidth]{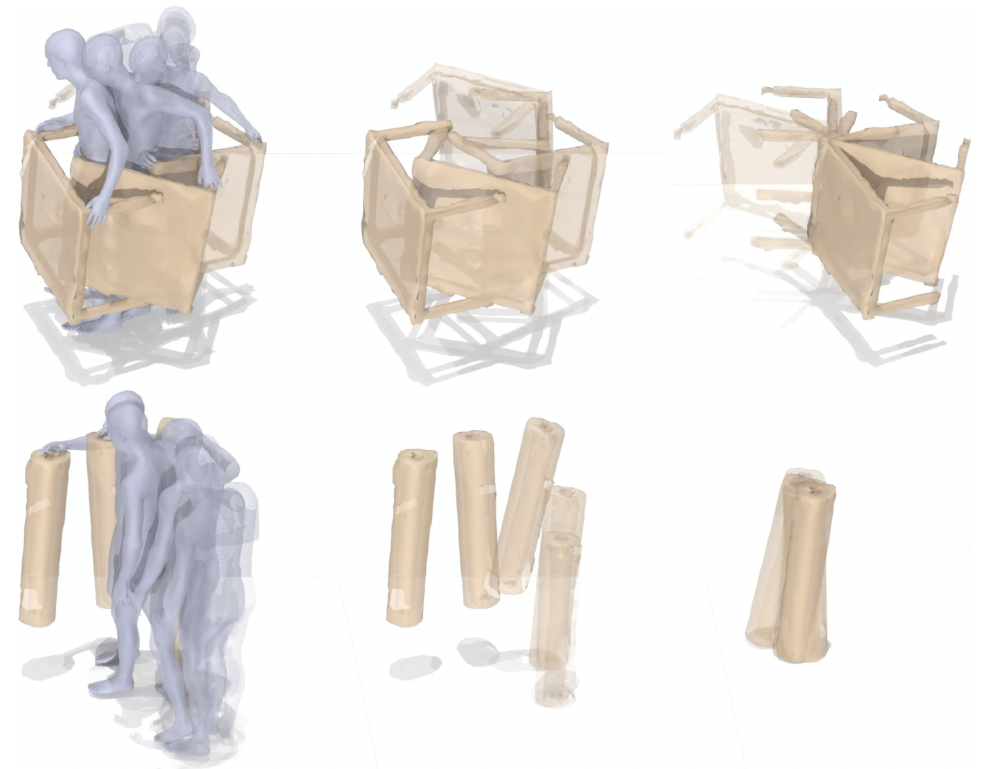}
	\vspace{-10pt}
	\caption{\small InterDiff~\cite{xu2023interdiff} applies coordinate transformations to represent object states relative to contact points, resulting in simpler motion patterns (right column) compared to using absolute positions (middle column). Figures are adapted from~\cite{xu2023interdiff}.}
	\label{fig:ho_spc_interdiff}
	\vspace{-10pt}
\end{figure}

\begin{figure*}
	\begin{center}
		\includegraphics[width=0.95\linewidth]{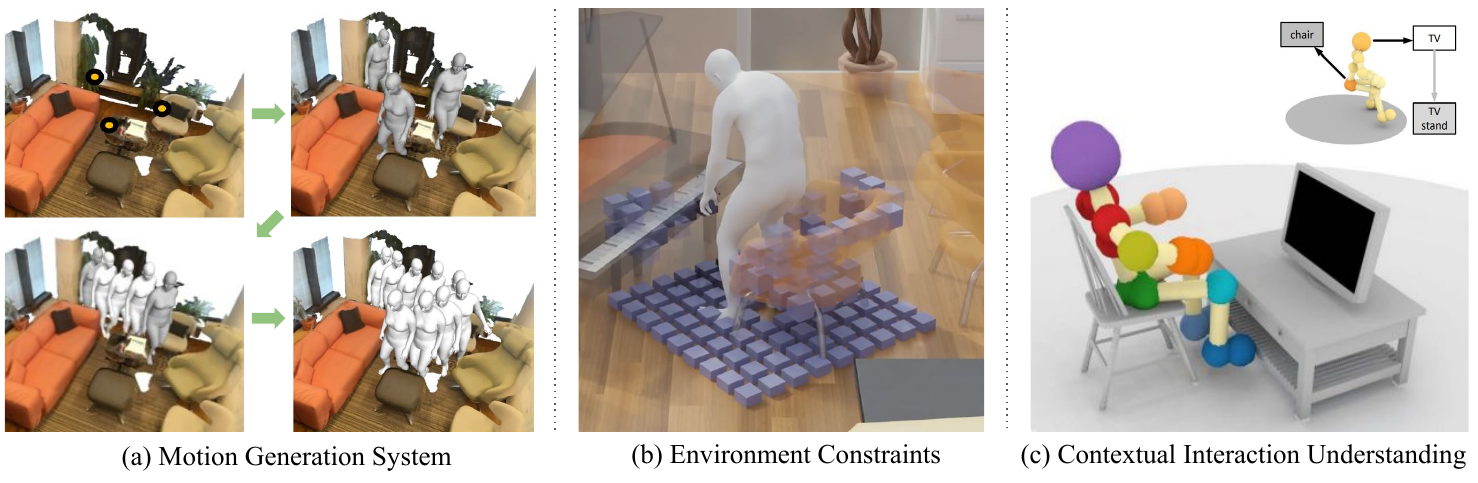}
	\end{center}
	\vspace{-5pt}
	\captionsetup{font=small}
	\caption{Existing works model human scene interactions in three major aspects: (a) \textit{Motion generation system} that decompose complex interactions into modular subtasks for systematic processing; (b) \textit{Environment constraint models} that incorporate physical constraints between humans and their surroundings; and (c) \textit{Contextual interaction understanding} that analyzes spatial relationships within the environment. Figures are adapted from~\cite{wanglongterm, LINGO, savva2016pigraphs}.}
	\label{fig:hs_challenges_viz}
	\vspace{-10pt}
\end{figure*}

\begin{figure}[t]
	\centering
	\includegraphics[width=0.48\textwidth]{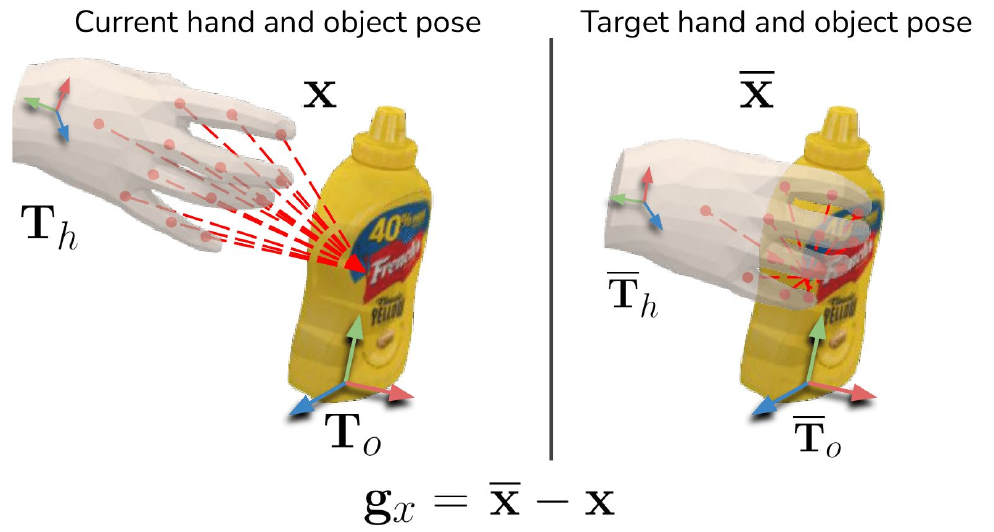}
	\vspace{-10pt}
	\caption{\small Grasp~\cite{christen2022dgrasp} introduces the target distance component \( g_x \) which computes the displacement between current and target 3D joint positions relative to the object's origin. Figures are adapted from~\cite{christen2022dgrasp}.}
	\label{fig:ho_spc_dgrasp}
	\vspace{-10pt}
\end{figure}

Contact-guided methods focus on optimizing the spatial relationships between hands and objects through contact points or regions. CHOIS~\cite{li2025controllable} and DiffH2O~\cite{christen2024diffh2o} implement contact guidance functions to minimize gaps and penetrations between the hand and the object meshes. Similarly, GraspDiff\cite{zuo2024graspdiff}, GRIP\cite{taheri2024grip}, and GrabNet~\cite{taheri2020grab} employ refinement networks utilizing Chamfer distance~\cite{bcd2021wu} and proximity sensors to enhance grasp quality. In the realm of contact optimization, ContactOpt~\cite{Grady_2021_CVPR} predicts contact points and optimizes hand poses to achieve desired contact configurations, while ContactGrasp~\cite{brahmbhatt2019contactgrasp} introduces an optimization framework that leverages attractive and repulsive values assigned to object surface points. ContactGen~\cite{Liu_2023_ICCV} advances these approaches by utilizing object-centric contact maps combined with~\textit{part maps} and~\textit{direction maps} to precisely locate hand contact points. RegionGrasp~\cite{wang2024regiongrasp} aims to generate hand grasp poses with the thumb finger contacting specific regions of the object. Recent diffusion-based methods such as HOI-Diff\cite{peng2023hoi} and CG-HOI\cite{diller2023cghoi} implement HOI affordance guidance during the sampling process to achieve coherent interaction synthesis. InterDiff~\cite{xu2023interdiff} suggests that object motion relative to contact points on the human body shows simpler learnable patterns than its absolute positions, as illustrated in Fig.~\ref{fig:ho_spc_interdiff}. In addition, DGrasp~\cite{christen2022dgrasp} incorporates the target distance as an input feature, promoting grasping at contact points proximal to those specified in the grasp reference data, as illustrated in Fig.~\ref{fig:ho_spc_dgrasp}.

Physics-based approaches have significantly improved interaction realism through simulation and learning. PhysFullbody\cite{braun2023physically} achieves full-body dexterous grasping through reinforcement learning in a physics simulation. ManipNet~\cite{zhang2021manipnet} demonstrates fine-grained synchronization through learned contact constraints, while DGrasp~\cite{christen2022dgrasp} implements a PD-controller in a physics simulation to compute hand torques for precise object manipulation. Extending these capabilities, ArtiGrasp~\cite{zhang2024artigrasp} synthesizes physically plausible bimanual grasping through reinforcement learning.

In addition, several innovative approaches have emerged for specialized interaction scenarios. NIFTY~\cite{kulkarni2024nifty} develops neural interaction fields attached to objects, guiding an object-conditioned human motion diffusion model to generate plausible contacts for diverse actions. GraspingField~\cite{GraspingField:3DV:2020} employs an implicit representation by encoding grasps in a signed distance field, enabling 3D hand-object grasp reconstruction from single images. For specific manipulation tasks, Chen et al.~\cite{chen2023synthesizing} focus on nonprehensile pre-grasp motions by optimizing thumb and index finger contact trajectories. GEARS~\cite{zhou2024gears} introduces a novel joint-centered point-based geometry sensor combined with spatiotemporal self-attention to model joint correlations effectively. G-HOP~\cite{ye2023ghop} advances the field by utilizing interaction grids with diffusion-based generative priors. Most recently, GeneOHDiffusion~\cite{liu2024geneoh} proposes a contact-centric HOI representation that parameterizes interactions in a region-specific coordinate system, significantly enhancing generalization across diverse interaction scenarios.

\vspace{-6pt}
\subsection{Human-Scene Interaction Generation}
\label{sec:interactive-hs-mg}
\begin{table*}[t]
	\centering
	\captionsetup{font=small}
	\caption{Representative works of human-scene interaction motion generation.}
	\label{table:hsmg_methods}
	\vspace{-5pt}
	\begin{threeparttable}
	\resizebox{0.99\textwidth}{!}{
	\setlength\tabcolsep{2.5pt}
	\renewcommand\arraystretch{1.02}
	\begin{tabular}{r||c|c|c|c|c|c}
	\hline\thickhline
	\rowcolor{mygray1}
        Method & Year & Venue & Model & Condition & Human-Scene Interaction Dataset & Other Human Dataset\\
	\hline
	\hline
         CLoSD~\cite{tevet2024closd}                                         &  2025 & ICLR          & RL, Physics Simulation   & Text Description          & -     &  AMASS~\cite{AMASS} \\
         GHOST~\cite{milacski2024ghostgroundedhumanmotion}                   &  2025 & WACV           & cVAE                    & Text Description          & HUMANISE~\cite{humanise}     &  - \\
         LINGO~\cite{LINGO}                                                  &  2024 & TOG            &  Diffusion              & Text Description          & LINGO~\cite{LINGO}           &  - \\
         UniHSI~\cite{unihsi}                                                &  2024 & ICLR           &  RL, Physics Simulation & Text Description          & ScenePlan~\cite{unihsi}      &  - \\
         TeSMo~\cite{tesmo}                                                  &  2024 & ECCV          &  Diffusion              & Text Description          & Loco-3DFRONT~\cite{tesmo}, SAMP~\cite{hassan2021stochastic}   & HumanML3D~\cite{HumanML3D} \\ 
         MOB~\cite{liu2023revisit}                                           &  2024 & ECCV               & cVAE               & Past Motion, Target Pose/Traj. & CIRCLE~\cite{CIRCLE}    & HumanML3D~\cite{HumanML3D} \\
         Xing et al.~\cite{xing2024sceneawarehumanmotionforecasting}         &  2024 & ECCV           & Graph Convolutional Network & Past Motion           & GTA-IM~\cite{caoHMP2020}, PROX~\cite{PROX}, HUMANISE~\cite{humanise}, GIMO~\cite{gimo} &  \\
         AffordMotion~\cite{AffordMotion}                                    &  2024 & CVPR              &  Diffusion              & Text Description          & HUMANISE~\cite{humanise}     & HumanML3D~\cite{HumanML3D} \\
         Cen et al.~\cite{cen2024text_scene_motion}                          &  2024 & CVPR              & Diffusion               & Text Description          & HUMANISE~\cite{humanise}     &  - \\
         GenZI~\cite{genzi}                                                  &  2024 & CVPR              & Optimization            & Text Description          & -           &  - \\
         Lou et al.~\cite{lou2024multimodal}                                 &  2024 & CVPR             & cVAE                    & Past Motion, Gaze         & GTA-IM~\cite{caoHMP2020}, GIMO~\cite{gimo} & -  \\
         S2Fusion~\cite{tang2024unified}                                     &  2024 & CVPR               & Diffusion               & Sparse Tracking Signals   & CIRCLE~\cite{CIRCLE}, GIMO~\cite{gimo} & -  \\
         TRUMANS~\cite{TRUMANS}                                              &  2024 & CVPR         &  Diffusion              & Action class              & TRUMANS~\cite{TRUMANS}       &  - \\
         InterScene~\cite{pan2023synthesizing}                               &  2024 & 3DV               &  RL, Physics Simulation & Object-Level Action Class       &     SAMP~\cite{hassan2021stochastic}               &  - \\
         Purposer~\cite{ugrinovic2024purposer}                               &  2024 & 3DV           & cVAE                      & Past Motion, Target Pose/Traj. & HUMANISE~\cite{humanise}  & BABEL~\cite{BABEL} \\
         MCLD~\cite{gao2024multiconditionlatentdiffusionnetwork}             &  2024 & IEEE TIP          & Diffusion               & Past Motion               & GTA-IM~\cite{caoHMP2020}, PROX~\cite{PROX} &  - \\  
         SIMS~\cite{wang2024simssimulatinghumansceneinteractions}            &  2024 &   -                & RL, Physics Simulation  & Text Script               & CIRCLE~\cite{CIRCLE}, SAMP~\cite{hassan2021stochastic}         & 100Style~\cite{mason2022realtimestylemodellinghuman} \\
         DIP~\cite{gong2024dip}                                              &  2024 &    -          & Diffusion               & Text Description          &                 -            & BABEL~\cite{BABEL}, HumanML3D~\cite{HumanML3D} \\
         GPT-Connect~\cite{gptconnect}                                       &  2024 &   -                &  Diffusion              & Text Description          & HUMANISE~\cite{humanise}     &  - \\
         LaserHuman~\cite{cong2024laserhuman}                                &  2024 & -                 & Diffusion               & Text Description          & LaserHuman~\cite{cong2024laserhuman} &  - \\    
         PAAK~\cite{mullen2023paak}                                          &  2023 & WACV             & Optimization            &            -              & PROX~\cite{PROX}             &  - \\
         DIMOS~\cite{DIMOS}                                                  &  2023 & ICCV              & RL                      & Target Pose               &  SAMP~\cite{hassan2021stochastic}     & AMASS~\cite{AMASS} \\
         LAMA~\cite{lama}                                                    &  2023 & ICCV               &  RL                     & Interaction Cue           &             -                & Self-Captured \\
         Narrator~\cite{Narrator}                                            &  2023 & ICCV               & cVAE                       & Text Description          & PROX~\cite{PROX}             &  - \\
         MAMMOS~\cite{mammos}                                                &  2023 & ICCVW             & cVAE                      & Action Class              & PROX~\cite{PROX}             & -  \\
         Hassan et al.~\cite{synthesizingphysicalcharactersceneinteractions} &  2023 & SIGGRAPH           & RL, Physics Simulation  & Tasks                     &   SAMP~\cite{hassan2021stochastic}   & - \\
         CIRCLE~\cite{CIRCLE}                                                &  2023 & CVPR                & Transformer             & Goal Position             & CIRCLE~\cite{CIRCLE}         &  - \\
         SceneDiffuser~\cite{SceneDiffuser}                                  &  2023 & CVPR              & Diffusion               &       -                  & PROX-S~\cite{zhao2022compositional}, LEMO~\cite{zhang2021learning} &  - \\
         Mir et al.~\cite{mir2023generatingcontinualhumanmotion}             &  2023 & 3DV               & Transformer             & Action Keypoints          &        -                    & AMASS~\cite{AMASS} \\
         HUMANISE~\cite{humanise}                                            &  2022 & NeurIPS            & cVAE                        & Text Description          & HUMANISE~\cite{humanise}     &  - \\
         COINS~\cite{zhao2022compositional}                                  &  2022 & ECCV              & cVAE                      & Object-Level Action Class & PROX-S~\cite{zhao2022compositional} &  - \\
         GAMMA~\cite{wanderings}                                             &  2022 & CVPR              & cVAE                     & Past Motion, Goal         &       -                      & AMASS~\cite{AMASS}  \\
         Wang et al.~\cite{Wang2022TowardsDA}                                &  2022 & CVPR               & cVAE                       & Target Actions            & PROX~\cite{PROX}             & -  \\
         SAMP~\cite{hassan2021stochastic}                                    &  2021 & ICCV               & cVAE                      & Target Goal            & SAMP~\cite{hassan2021stochastic}              & -  \\
         Wang et al.~\cite{wanglongterm}                                     &  2021 & CVPR               & cVAE                     & Body Sub-Goals            & PROX~\cite{PROX}             &  - \\
         Cao et al.~\cite{caoHMP2020}                                        &  2020 & ECCV              & cVAE                        & Past Motion               & GTA-IM~\cite{caoHMP2020}          &  - \\   
         PSI~\cite{PSI:2019}                                                 &  2020 & CVPR               & cVAE                       &             -             & PROX~\cite{PROX}             &  - \\
         PLACE~\cite{zhang2020place}                                         &  2020 & 3DV                & cVAE, Optimization               &           -               & PROX~\cite{PROX}             & -  \\
         Neural State Machine~\cite{NSM}                                     &  2019 & TOG            & RL                      & Goal Action               &             -                & Self-Captured \\
         3DAffordance~\cite{3daffordance}                                    &  2019 & CVPR               &   GAN, Optimization       &      Scene Image            & -                             & -  \\
         PiGraphs~\cite{savva2016pigraphs}                                   &  2016 & TOG              & Interaction Graph                        & Object Relationships      & PiGraphs~\cite{savva2016pigraphs}  & - \\
	\hline
	\end{tabular}
	}
	\end{threeparttable}
	\vspace{-5pt}
\end{table*}

In everyday life, humans effortlessly navigate and interact with complex environments. Recreating this natural ability is essential for 3D applications such as gaming, simulation, and character animation in virtual environments. Human-scene interaction (HSI) entails modeling how humans move and interact with their surroundings while adhering to a set of physical rules, such as collision avoidance and aligning with semantic or contextual constraints. Additionally, supplementary signals—such as text prompts, action labels, and target goals—further enhance the control and precision of motion generation in these systems.

Existing work attempts to achieve scene-aware motion generation through these three pillars: motion generation systems, environment constraints, and contextual interaction understanding, as illustrated in Fig.~\ref{fig:hs_challenges_viz}.
\textit{Motion generation system} decomposes the complex scene-aware motion synthesis into modular and more tractable subtasks, a common strategy in recent researches.
\textit{Environment constraints} ensure physical plausibility by maintaining coherence between generated motions and spatial constraints. These constraints align human poses and movements with scene geometry and surfaces, preserving natural motion dynamics and physical consistency throughout the interaction sequence.
\textit{Contextual interaction understanding} enhances system capabilities through semantic comprehension of environmental context, enabling meaningful and nuanced interactions. This encompasses graph-based modeling of object relationships for joint-level interactions, high-level planning through LLMs, and interpretation of scene affordances via image-based cues.

These three aspects are fundamental to generating lifelike, physically coherent, and semantically appropriate human motions in 3D scenes, forming the foundation for contemporary advances in scene-aware motion generation. We summarize the characteristics of human-scene interaction generation methods in Table~\ref{table:hsmg_methods}. The following sections explore recent advancements in addressing these challenges within scene-aware motion generation.

\subsubsection{Building Motion Generation System}

Several methods adopt hierarchical or stage-wise pipelines to synthesize interactions with the scene. 
COINS~\cite{zhao2022compositional} first generates a plausible human pelvis pose and then completes the body pose for static human-scene interaction.
Wang et al.~\cite{wanglongterm} propose a framework for synthesizing long-term motion by generating static sub-goal poses from given root information and connecting them with a short-term motion completion module.
SAMP~\cite{hassan2021stochastic} first predicts an oriented goal location for interaction, then generates a collision-free path, and finally predicts the full-body motion required to reach the target.
Wang et al.~\cite{Wang2022TowardsDA} first synthesize human-scene interaction anchors based on action labels, then complete motions by integrating these anchors with planned paths.
PAAK~\cite{mullen2023paak} proposes a method for placing existing human animations into scenes by identifying keyframes that are most critical for interactions with the scene and using them to accurately position the animation within the environment.
MAMMOS~\cite{mammos} synthesizes interaction anchors for each individual character and completes motions by integrating timelines to account for multi-human interactions.
Mir et al.~\cite{mir2023generatingcontinualhumanmotion} first infer static key poses from action keypoints defined by users and then complete a variety of motion types—such as transitions in/out and walking—utilizing a goal-centric canonical coordinate space to complete motion sequences.
TeSMo~\cite{tesmo} synthesizes navigation motion based on the generated root trajectory and subsequently produces interaction motion with the target object.
DIP~\cite{gong2024dip} decomposes the entire command task into a list of subtasks and identifies corresponding keyframes. Then it completes motions conditioned on actions and keyframe joints, which are subsequently refined through time-variant blending.

The auto-regressive motion generation strategy enables the generation of arbitrarily long sequences, and several methods adopt this approach.
MOB~\cite{liu2023revisit} formulates an auto-regressive motion controller by utilizing historical motion states and integrating target positions or trajectories, demonstrating its capability to produce motion in both static and dynamically changing scenes.
MCLD~\cite{gao2024multiconditionlatentdiffusionnetwork} predicts future human motion from past body movements using a conditional diffusion formulation.
TRUMANS~\cite{TRUMANS} and LINGO~\cite{LINGO} employ an auto-regressive diffusion sampling approach to enable the progressive generation of long motion sequences. They replace the preceding segment of frames with the previous corresponding last segment and then generate continuous motion from these.

Another line of work formulates motion generation as a Markov decision process and utilizes reinforcement learning (RL) based frameworks to generate continuous motion sequences.
GAMMA~\cite{wanderings} uses a policy network to explore latent representations for marker prediction and applies a tree-based search to find motion primitives.
DIMOS~\cite{DIMOS} creates static human-scene interaction poses for sub-goals and uses a policy network to generate scene-conditioned latent actions, which are later converted into motion primitives by a pre-trained model.
LAMA~\cite{lama} derives action signals from a motion controller, transforms them into motion features, and identifies motion segments using motion-matching algorithms. These motion sequences are further optimized within a learned manifold for scene manipulation.
Additionally, physics simulators, such as~\cite{makoviychuk2021isaac}, have been applied to this formulation.
Hassan et al.~\cite{synthesizingphysicalcharactersceneinteractions} train a policy and motion discriminator to achieve goal-based interactions with the scene and generate natural motion consistent with the dataset.
UniHSI~\cite{unihsi} transforms language commands into an ordered condition for interaction, referred to as the Chain of Contacts (CoC), and designs a robust unified controller for executing various interaction tasks.
InterScene~\cite{pan2023synthesizing} employs a finite-state machine to coordinate navigation and interaction controllers, enabling long-term interactions within a scene.
SIMS~\cite{wang2024simssimulatinghumansceneinteractions} combines high-level LLM planning with low-level physical control by extracting interactions from real videos into short scripts, which are then combined into long-term narratives. The system matches these narratives with suitable scene layouts using graph-based methods and also applies a finite-state machine to translate scene information into physical control policies for practical execution.

\subsubsection{Capturing Environment Constraints}
A key challenge in scene-aware motion generation is maintaining adherence to spatial constraints and physical rules. One approach uses post-optimization or physics simulation techniques to refine the generated motion, ensuring that it aligns with these constraints and rules.
PSI~\cite{PSI:2019} refines the static body results to encourage contact and prevent interpenetration between the body and the scene.
COINS~\cite{zhao2022compositional} refines the static pose results to enhance physical plausibility by aligning predicted contact regions with objects according to the specified actions.
Wang et al.~\cite{wanglongterm} leverage a cVAE to generate initial static poses conditioned on 3D scenes. After generating the full motion sequence from anchors, they refine the resulting motions through post-optimization, mitigating collisions and enhancing contact at body vertices, as well as motion smoothness and foot constraint.
Wang et al.~\cite{Wang2022TowardsDA} introduce Neural Mapper to generate diverse, obstacle-free paths that extend conventional path-finding algorithms. Building upon a path, they complete motion and further optimize it by utilizing a physical loss.
Narrator~\cite{Narrator} introduces an Interaction Bisector Surface (IBS) loss to effectively handle penetration and contact issues during scene-aware optimization. 
CLoSD~\cite{tevet2024closd} enhances physical plausibility by incorporating a physics simulator after diffusion-based generation.

\begin{table*}[t]
	\centering
	\captionsetup{font=small}
	\caption{Representative works of human-mix interaction motion generation. (HH: Human-Human, HO: Human-Object, HS: Human-Scene.)}
	\label{table:hmmg_methods}
	\vspace{-5pt}
	\begin{threeparttable}
	\resizebox{0.99\textwidth}{!}{
	\setlength\tabcolsep{2.5pt}
	\renewcommand\arraystretch{1.02}
	\begin{tabular}{r||c|c|c|c|c|c}
	\hline\thickhline
	\rowcolor{mygray1}
        Method & Year & Venue & Tasks & Model & Condition & Dataset\\
	\hline
	\hline
        Sitcom-Crafter~\cite{chen2024sitcomcrafterplotdrivenhumanmotion} & 2025 & ICLR & HH, HS & Diffusion & Scene, Text Description & InterHuman~\cite{InterGen}, Inter-X~\cite{xu2023interx} \\
        HOI-M\(^3\)~\cite{zhang2024hoim3} & 2024 & CVPR & HH, HO & Diffusion & Object & HOI-M\(^3\)~\cite{zhang2024hoim3} \\
        Shu et al.~\cite{shu2016HHOI} & 2016 & IJCAI & HH, HO & Graph & Past Motion, Object & HHOI~\cite{shu2016HHOI} \\
	\hline
	\end{tabular}
	}
	\end{threeparttable}
	\vspace{-5pt}
\end{table*}

Another approach focuses on effectively modeling spatial constraints during the process of motion generation. These methods can be categorized based on their key technical contributions:
First, several works focus on encoding scene-body relationships through specific feature representation. PLACE~\cite{zhang2020place} explicitly models body-environment proximity using BPS~\cite{prokudin2019efficient} within a cVAE architecture. MOB~\cite{liu2023revisit} learns motion-scene relationships through a canonicalized space occupancy grid, training exclusively on human-only motion data. AffordMotion~\cite{AffordMotion} introduces an affordance map based on skeleton-scene distance fields, generating motions from language instructions via this intermediate representation.  S2Fusion~\cite{tang2024unified} combines PointNet-extracted scene features with periodic tracking signals. Xing et al.\cite{xing2024sceneawarehumanmotionforecasting} propose a mutual distance representation incorporating per-vertex signed distances and per-basis point distances. TRUMANS~\cite{TRUMANS} conditions a diffusion model on local occupancy voxels centered on the pelvis, encoded via ViT~\cite{dosovitskiy2021imageworth16x16words}. On top of this, a few other works focus on novel architectural designs and learning strategies. CIRCLE~\cite{CIRCLE} employs a transformer-based refinement module utilizing scene geometry features from BPS or PointNet~\cite{qi2017pointnet}, derived from initial body surfaces. SceneDiffuser~\cite{SceneDiffuser} incorporates a cross-attention module for scene point cloud encoding and integrates physics-based objectives during sampling to ensure proper contact and collision avoidance.  DIP~\cite{gong2024dip} enhances the diffusion process with implicit policy optimization, guided by contact and non-penetration rewards. Lou et al.\cite{lou2024multimodal} extract both local and global salient points for detailed spatial understanding and trajectory planning, integrated through scene-aware cross-modal attention. LINGO~\cite{LINGO} employs dual voxels for current and predictive contexts, with adaptive positioning based on motion stage and interaction targets.

\subsubsection{Understanding Spatial Interaction Context}
An advanced understanding of the spatial layout and relationships in the scene is also important for HSI generation. This information would navigate the character in the scene and direct the interactions with specific scene entities.

PiGraphs~\cite{savva2016pigraphs} learns a probabilistic graph of human-centric interaction, where nodes represent objects and body parts, and edges encode the spatial relationships between interacting joints and objects. Purposer~\cite{ugrinovic2024purposer} proposes a method for estimating dense body-to-scene contact using vertex-level semantic labels, enabling the semantically appropriate placement of posed humans in 3D scenes. Neural State Machine~\cite{NSM} extracts extensive motion features regarding its surrounding cylinder environment, and employs a gated network and a motion prediction network that processes the current state and the goal action vector to predict posture and control variables for sequential frames. COINS~\cite{zhao2022compositional} synthesizes static human-scene interactions derived from `action-object' semantics, and combines atomic interactions into compositional interactions.
Narrator~\cite{Narrator} employs a \textit{joint global and local scene graph} for spatial relationship modeling and a \textit{part-level action} mechanism to align body parts with actions for realistic and text-faithful interactions.

Recent advances in text-conditioned scene-aware motion generation have leveraged both semantic understanding and large language models (LLMs). GHOST~\cite{milacski2024ghostgroundedhumanmotion} enhances motion synthesis by learning a shared text-scene feature space that improves semantic scene understanding. GPT-Connect~\cite{gptconnect} utilizes LLMs to comprehend scene context and generate informative human skeleton configurations, which subsequently guide the motion diffusion model's output. Taking a unified approach, UniHSI~\cite{unihsi} formulates human-scene interaction through sequential transitions of joint-object contact pairs, employing an LLM planner to transform language commands into structured task plans. Building on this trajectory, Cen et al.~\cite{cen2024text_scene_motion} integrate scene graph construction with LLM-based textual analysis to identify target objects and synthesize contextually appropriate interaction motions.

Another line of work utilizes image information to guide the motion generation process.
3DAffordance~\cite{3daffordance} generates semantically plausible poses on scene images and maps human poses into scene voxels.
Cao et al.~\cite{caoHMP2020} predict human motion influenced by the scene using a single-scene image and 2D pose histories as input.
GenZI~\cite{genzi} synthesizes 2D human interactions in image space using Vision-Language Models (VLMs) with a textured 3D scene across multiple views, lifting them to 3D through a robust optimization process for human parameters.

\vspace{-6pt}
\subsection{Human-Mix Interaction Generation}
\label{sec:interactive-hm-mg}
Recent research has expanded to address more complex environments involving multiple entities—objects, scenes, and other humans—in interactive motion synthesis. We summarize the characteristics of human-mix interaction generation methods in Table~\ref{table:hmmg_methods}.

Shu et al.\cite{shu2016HHOI} pioneer the exploration of human-human-object interactions by reducing the problem complexity through object trajectory abstraction. By treating objects as additional joints, their approach leverages graph models to predict interactive human motions, enabling simultaneous modeling of human-human and human-object close interactions. HOI-M3\cite{zhang2024hoim3} advances this framework by unifying multi-human information and multi-object poses into a comprehensive motion embedding. The system employs PointNet~\cite{qi2017pointnet} for extracting geometric object features and utilizes a diffusion model with concatenated object and motion embeddings to generate semantically coherent interactive motions for both humans and objects.

\begin{table*}[t]
	\centering
	\captionsetup{font=small}
	\caption{\small{\textbf{Human-human interaction datasets.} This table summarizes key statistics and features of various human-human interaction datasets. Subjects: The number of individuals involved in the dataset; Sequences: The number of motion clips available; Frames: The total number of frames capturing 3D human motions; Length: The cumulative duration of the dataset's motion data (in hours); Acquisition: The method used to obtain motion data  (e.g., multi-view RGB videos denoted as ``mRGB''); Modality: The representation format of motion data; Video, Text, Audio: Indicates whether the dataset includes corresponding modalities.}}
	\label{table:hhmg_datasets}
	\vspace{-5pt}
	\begin{threeparttable}
	\resizebox{0.99\textwidth}{!}{
	\setlength\tabcolsep{2.5pt}
	\renewcommand\arraystretch{1.02}
	\begin{tabular}{r||c|c|c|c|c|c|c|c|c|c|c}
	\hline\thickhline
	\rowcolor{mygray1}
        Dataset & Year & Venue & Subjects & Sequences & Frames & Length & Acquisition & Modality & Video & Text & Audio \\
	\hline
	\hline
        Inter-X~\cite{xu2023interx} & 2024 & CVPR & 89 & 11,388 & 8.1M & - & MoCap & SMPL-X & \xmark & \cmark & \xmark \\
        InterHuman~\cite{InterGen} & 2024 & IJCV & 60 & 7,779 & 107M & 6.56h & mRGB & SMPL & \xmark & \cmark & \xmark \\
        ReMoCap~\cite{ghosh2024remos} & 2024 & ECCV & 9 & - & 275.7K & 2.04h & mRGB & 3D Skeleton & \cmark & \xmark & \xmark \\
        LAION-Pose (image-only)~\cite{shan2024opendomain} & 2024 & ECCV & - & - & 8M & - & sRGB & SMPL & \xmark & \cmark & \xmark \\
        WebVid-Motion~\cite{shan2024opendomain} & 2024 & ECCV & - & 3,500 & - & - & sRGB & SMPL & \cmark & \cmark & \xmark \\
        DD100~\cite{siyao2024duolando} & 2024 & ICLR & 10 & 100 & 210K & 1.95h & MoCap & SMPL-X & \xmark & \xmark & \cmark \\
        Hi4D~\cite{yin2023hi4d} & 2023 & CVPR & 40 & 100 & 11K & - & mRGB & SMPL & \cmark & \xmark & \xmark \\
        GTA Combat~\cite{xu2022actformer} & 2023 & ICCV & 14 & 6,900 & - & - & MoCap (GTA-V) & 3D Skeleton & \cmark & \xmark & \xmark \\
        ExPI~\cite{guo2022multipersonexmotion} & 2022 & CVPR & 4 & 115 & 30K & 0.33h & MoCap + mRGB & 3D Skeleton + 3D Mesh & \cmark & \xmark & \xmark \\
        MultiHuman~\cite{zheng2021deepmulticap} & 2021 & ICCV & 278 & 150 & - & - & mRGB & SMPL-X & \cmark & \xmark & \xmark \\
        Chi3D~\cite{Fieraru_2020_CVPR} & 2020 & CVPR & 6 & 631 & 486K & 2.70h & MoCap + mRGB & GHUM + SMPL-X & \cmark & \xmark & \xmark \\
        You2Me~\cite{ng2020you2me} & 2020 & CVPR & 10 & 42 & 77K & 1.4h & mRGB-D & 3D Skeleton & \cmark & \xmark & \xmark \\
        2C~\cite{2C} & 2020 & TVCG & 2 & 8 & - & 0.06h & MoCap & 3D Skeleton & \xmark & \xmark & \xmark \\
        DuetDance~\cite{Kundu_2020_WACV} & 2020 & WACV & - & - & 196K & 1.09h & - & - & \xmark & - & - \\
        NTU RGB+D 120 (interactive)~\cite{Liu_2020_NTURGBD120} & 2019 & TPAMI & 106 & 20,579 & - & 18.6h & RGB-D & 3D Skeleton & \cmark & \xmark & \xmark \\
        Talking With Hands 16.2M~\cite{Lee_2019_ICCV} & 2019 & ICCV & 50 & 200 & 16.2M & 50h & MoCap & 3D Skeleton & \xmark & \xmark & \cmark \\
        Haggling~\cite{joo2019cvpr} & 2019 & CVPR & 102 & 34 & - & 6h & mRGB-D & 3D Skeleton & \cmark & \xmark & \cmark \\
        MuPoTS-3D~\cite{mehta2018ssmp} & 2018 & 3DV & 8 & 20 & 8K & - & mRGB & 3D Skeleton & \cmark & \xmark & \xmark \\
        3DPW~\cite{Marcard_2018_ECCV} & 2018 & ECCV & 18 & 60 & 51K & - & sRGB + IMU & SMPL & \cmark & \xmark & \xmark \\
        JTA~\cite{fabbri2018jta} & 2018 & ECCV & 10,800 & 512 & 461K & 4.27h & MoCap (GTA-V) & 3D Skeleton & \cmark & \xmark & \xmark \\
        CMU-Panoptic~\cite{joo2016panopticstudio} & 2017 & TPAMI & - & 65 & 1.5M & 5.5h & mRGB-D & 3D Skeleton & \cmark & \xmark & \cmark \\
        ShakeFive2~\cite{ShakeFive2} & 2016 & HBU & 10 & 153 & 34K & 0.32h & RGB-D & 3D Skeleton & \cmark & \xmark & \xmark \\
        MARCOnI~\cite{Elhayek_2015_CVPR} & 2015 & CVPR & 10 & 12 & 6.2K & 0.07h & mRGB & 3D Skeleton & \cmark & \xmark & \xmark \\
        K3HI~\cite{K3HI} & 2013 & Math. Probl. Eng. & 15 & 320 & 8K & - & RGB-D & 3D Skeleton & \xmark & \xmark & \xmark \\
        SBU Kinect~\cite{sbu_kinect} & 2012 & CVPRW & - & 300 & 7.5K & 0.13h & RGB-D & 3D Skeleton & \cmark & \xmark & \xmark \\
        UMPM~\cite{umpm} & 2011 & ICCVW & 30 & 36 & 400K & 2.22h & MoCap + mRGB & 3D Skeleton & \cmark & \xmark & \xmark \\
        MHHI~\cite{MHHI} & 2011 & CVPR & 5 & 7 & 1.5K & 0.01h & MoCap + mRGB & 3D Skeleton + 3D Mesh & \cmark & \xmark & \xmark \\
        UT-Interaction~\cite{UT-Interaction-Data} & 2010 & ICPR & 15 & 20 & 36K & 0.33h & sRGB & No Skeleton & \cmark & \xmark & \xmark \\
        CMU Graphics Lab MoCap (interactive)~\cite{cmu_mocap} & 2008 & - & 6 & 53 & 7.6K & 0.07h & MoCap + sRGB & 3D Skeleton & \cmark & \xmark & \xmark \\
	\hline
	\end{tabular}
	}
	\end{threeparttable}
	\vspace{-5pt}
\end{table*}

Sitcom-Crafter~\cite{chen2024sitcomcrafterplotdrivenhumanmotion} addresses the challenge of generating expressive human-human interactions within 3D scenes. Acknowledging the scarcity of datasets that combine human-human interactions with 3D scene information, the authors propose an innovative solution: synthesizing implicit 3D Signed Distance Function points to simulate spatial obstacles. This synthetic training data enables effective collision avoidance while preserving motion fidelity.

\section{Datasets}
\label{sec:datasets}
Over recent decades, researchers have developed multiple datasets for human interaction motion generation research. In this section, we present a comprehensive analysis of widely adopted datasets across four primary interaction categories: human-human (Table~\ref{table:hhmg_datasets}), human-object (Table~\ref{table:homg_datasets}), human-scene (Table~\ref{table:hsmg_datasets}), and human-mix (Table~\ref{table:hmmg_datasets}) interactions.

\vspace{-6pt}
\subsection{Human-Human Interaction Datasets}
\label{sec:hh-data}

\begin{table*}[t]
	\centering
	\captionsetup{font=small}
	\caption{\small{\textbf{Human-object interaction datasets.} This table summarizes key statistics and features of various human-object interaction datasets. Subjects: The number of individuals involved in the dataset; Sequences: The number of motion clips available; Frames: The total number of frames capturing 3D human motions; Length: The cumulative duration of the dataset's motion data (in hours); Acquisition: The method used to obtain motion data  (e.g., multi-view RGB videos denoted as ``mRGB''); Modality: The representation format of motion data; Images, Text: Indicates whether the dataset includes corresponding modalities.}}
	\label{table:homg_datasets}
	\vspace{-5pt}
	\begin{threeparttable}
	\resizebox{0.99\textwidth}{!}{
	\setlength\tabcolsep{2.5pt}
	\renewcommand\arraystretch{1.02}
	\begin{tabular}{r||c|c|c|c|c|c|c|c|c|c|c|c|c}
	\hline\thickhline
	\rowcolor{mygray1}
        Dataset & Year & Venue & Subjects & Object Types & Sequences & Frames & Length & Acquisition & Modality & Images & Text\\
	\hline
	\hline
        HIMO~\cite{lv2024himo} & 2024 & ECCV & $34$ & $53$ Household Objects & $3,376$ & $4.08$M & $9.44$h & MoCap & SMPL-X & \xmark &  \cmark \\
        Semantic-HOI~\cite{yang2024f} & 2024 & ECCV & - & - & - & $20.4$K & - & Existing Datasets + GPT-4V & SMPL-X & \cmark & \cmark \\
        IMHD2~\cite{zhao2024imhoi} & 2024 & CVPR & $15$ &$10$ Everyday Objects & $295$ & $892$K & - & mRGB + IMU Sensor & SMPL-H & \cmark &  \xmark \\
        OMOMO~\cite{li2023object} & 2023 &TOG & $17$ & $15$ & - & - & - & mRGB & SMPL-X & \xmark & \cmark \\
        CHAIRS~\cite{jiang2023full} & 2023 & ICCV & $46$ & $81$ Sittable Objects & $1,390$ & $1.8$M & $17.3$h & mRGB + MoCap & SMPL-X & \xmark &  \xmark  \\
        ARCTIC~\cite{fan2023arctic} & 2023 & CVPR & $10$ & $11$ Articulated Objects & $339$ & $2.1$M & $1.2$h & mRGB + MoCap & SMPL-X & \xmark & \xmark \\
        HODome~\cite{zhang2023neuraldome} & 2023 & CVPR & $10$ & $23$ & $274$ & $71$M & $4.5$h & mRGB & SMPL &\xmark & \xmark \\
        BEHAVE~\cite{bhatnagar2022behave} & 2022 & CVPR & $8$ & $20$ Common Objects & $321$ & $15.2$K & $4.2$h & mRGB-D & SMPL & \cmark &  \xmark \\
        HOI4D~\cite{Liu_2022_CVPR} & 2022 & CVPR & - & - & - & - & - & - & - & \xmark &  \xmark \\ 
        H2O3D~\cite{hampali2022keypoint} & 2022 & CVPR &  $5$ & $10$ & - & $75$K & - & mRGB-D & MANO & \cmark &  \xmark \\ 
        COUCH~\cite{zhang2022couch} & 2022 & ECCV & $6$ & $3$ Chairs & $500$ & $648$K & $3$h & MoCap & SMPL & \xmark &  \xmark\\
        InterCap~\cite{huang2022intercap} & 2022 & GCPR & $10$ & $10$ Everyday Objects & $223$ & $67.3$K & - & mRGB & SMPL-X & \cmark &  \xmark \\ 
        GRAVICAP~\cite{GraviCap2021} & 2021 & ICCV & $4$ & $4$ Different balls & $9$ & - & - & mRGB + MoCap & 3D Skeleton & \xmark &  \xmark \\
        H2O~\cite{Kwon_2021_ICCV} & 2021 & ICCV & $4$ & - & - & $571$K & - & mRGB-D & MANO & \cmark & \xmark \\ 
        DexYCB~\cite{chao2021dexycb} & 2021 & CVPR & $10$ & $20$& $1000$ & $582$K & - & mRGB-D & 3D Keypoints & \cmark & \xmark \\ 
        GRAB~\cite{taheri2020grab} & 2020 & ECCV & $10$ & $51$ Small Objects & $1334$  & $1.6$M & $3.8$h & MoCap & SMPL-X & \xmark &  \xmark \\
        ContactPose~\cite{brahmbhatt2020contactpose} & 2020 & ECCV & $50$ & $25$ & -& -&- & mRGB-D & MANO & \cmark &  \xmark \\
        HO-3D~\cite{hampali2020honnotate} & 2020 & CVPR & $10$ & $10$ & $68$ & $77.5$K & - & RGB-D & MANO & \cmark & \xmark \\
        ObMan~\cite{hasson19_obman} & 2019 & CVPR & $20$ & $8$ Everyday Objects & -& $21$K & -& GraspIt & SMPL-H &\cmark &  \xmark \\
        ContactDB~\cite{Brahmbhatt_2019_CVPR} & 2019 & CVPR & $50$ & $50$ Household Objects & - & $375$K & - & mRGB-D & 3D Mesh & \xmark &  \xmark \\
        Dexter+Object~\cite{RealtimeHO_ECCV2016} & 2016 & ECCV & $2$ & $2$ Cuboids & $6$ & $3$K & - & RGB-D & 2D Keypoints &\cmark &  \xmark \\

    \hline
	\end{tabular}
	}
	\end{threeparttable}
	\vspace{-5pt}
\end{table*}

We review existing HHI motion datasets based on their distinct representations and conditioning modalities. Table \ref{table:hhmg_datasets} provides detailed statistics of current HHI datasets.

\paragraph{Human Skeleton Based Datasets.} These datasets primarily represent human interactions through 3D skeletal data, capturing joint coordinates and body keypoints at sequential frames. Early contributions in this domain include CMU Graphics Lab MoCap~\cite{cmu_mocap}, UMPM~\cite{umpm}, SBU Kinect~\cite{sbu_kinect}, K3HI~\cite{K3HI}, MARCOnI~\cite{Elhayek_2015_CVPR}, ShakeFive2~\cite{ShakeFive2}, MuPoTS-3D~\cite{mehta2018ssmp}, and 2C~\cite{2C}. These pioneering datasets employ single RGB-D or multi-camera setups to capture multi-human interactions, providing sparse representations of human motion.
More recent datasets have significantly expanded in scope, encompassing larger subject pools, longer action sequences, and more diverse social interactions. The NTU RGB+D 120~\cite{Liu_2020_NTURGBD120} dataset stands as a prominent example, featuring over 20,000 diverse motion sequences that have made it a standard benchmark in the field. Specialized datasets have also emerged, such as ReMoCap~\cite{ghosh2024remos}, which focuses on specific interaction contexts like dance and martial arts, providing two hours of dedicated human interaction data.
In addition, innovative data acquisition methods have been purposed. You2Me~\cite{ng2020you2me} utilizes wearable devices to capture reactor motions from the actor's perspective, establishing an innovative methodology for interaction data acquisition. Taking advantage of gaming technology, GTA Combat~\cite{xu2022actformer} and JTA~\cite{fabbri2018jta} leverage the GTA-V engine to generate large-scale, visually consistent multi-person scenarios featuring diverse human poses and versatile multi-human interactions.

\paragraph{Human Shape Based Datasets.}
Beyond 3D skeleton data, several datasets incorporate 3D body mesh information, enabling high-fidelity modeling of human-human interactions through more comprehensive body structure representation. Datasets like 3DPW~\cite{Marcard_2018_ECCV}, Chi3D~\cite{Fieraru_2020_CVPR}, MultiHuman~\cite{zheng2021deepmulticap}, and Hi4D~\cite{yin2023hi4d} utilize parametric body models, such as SMPL~\cite{pavlakos2019smpl-x}, SMPL-X~\cite{smpl}, and GHUM~\cite{Xu_2020_CVPR_GHUM}, to capture subtle variations in full-body shape and pose. In contrast, some datasets, like MHHI~\cite{MHHI} and ExPI~\cite{guo2022multipersonexmotion}, adopt a different approach by constructing custom 3D human meshes from captured motion actors, achieving more realistic motion representation and better alignment between 3D meshes and visual data.

\paragraph{Text-Conditioned Human-human Datasets.}
Recent advances in multi-human interaction datasets have incorporated textual descriptions. Pioneering datasets like InterX~\cite{xu2023interx} and InterHuman~\cite{InterGen} feature human-generated natural language annotations that provide detailed descriptions of both actor and reactor movements at the body part level for specific interaction sequences. In contrast, LAION-Pose~\cite{shan2024opendomain} and WebVid-Motion~\cite{shan2024opendomain} take an automated approach, utilizing Visual Question Answering (VQA) through the InstructBLIP~\cite{dai2023instructblip} model to generate captions. These automated annotations are obtained by prompting the model to "describe the person or group of people's action and body poses in the image."

\paragraph{Audio-Conditioned Human-human Datasets.}
Several datasets extend beyond visual and textual modalities by integrating audio signals to enrich multimodal analysis. CMU-Panoptic~\cite{joo2016panopticstudio} and Haggling~\cite{joo2019cvpr} employ the same multi-camera system to capture synchronized audiovisual data. Haggling is specifically designed to record three-person conversations, enabling the study of social dynamics in negotiation scenarios, whereas CMU-Panoptic expands its scope to include instrumentally performed music, background music, and noise generated by human motion. Talking With Hands 16.2M~\cite{Lee_2019_ICCV} emphasizes non-verbal communication by synchronizing conversational audio with detailed annotations of hand and body movements, supporting research on gesture and interaction analysis. Furthermore, DD100~\cite{siyao2024duolando} includes approximately two hours of professional dance performances, which capture various dance movements in different genres of background music, such as Samba and Tango.

\vspace{-6pt}
\subsection{Human-Object Interaction Datasets}
\label{sec:ho-data}
Recent years have witnessed the emergence of diverse datasets specifically designed for 3D human-object interaction modeling, each addressing distinct aspects of interaction complexity. Table \ref{table:homg_datasets} presents comprehensive statistics of these human-object interaction datasets.

\begin{table*}[t]
	\centering
	\captionsetup{font=small}
	\caption{\small{\textbf{Human-scene interaction datasets.} This table summarizes key statistics and features of various human-scene interaction datasets. Subjects: The number of individuals involved in the dataset; Sequences: The number of motion clips available; Frames: The total number of frames capturing 3D human motions; Length: The cumulative duration of the dataset's motion data (in hours); Motion Acquisition: The method used to obtain motion data; Scene Acquisition: The method used to obtain scene data; Modality: The representation format of motion data; Dynamic: Indicates whether the dataset includes dynamic or static scene.}}
	\label{table:hsmg_datasets}
	\vspace{-5pt}
	\begin{threeparttable}
	\resizebox{0.99\textwidth}{!}{
	\setlength\tabcolsep{2.5pt}
	\renewcommand\arraystretch{1.02}
	\begin{tabular}{r||c|c|c|c|c|c|c|c|c|c|c|c|c}
	\hline\thickhline
	\rowcolor{mygray1}
        Dataset & Year & Venue & Subjects & Scenes & Sequences & Frames & Length & Motion Acquisition & Scene Acquisition & Modality & Condition & Video & Dynamic \\
	\hline
	\hline
        LINGO~\cite{LINGO}                 & 2024 & TOG         &  - & 120 &   -   &   -   & 16h  & MoCap & VR Environment & SMPL-X & Text & \cmark &  \cmark \\
        TRUMANS~\cite{TRUMANS}             & 2024 & CVPR          &  5 & 100 &  -   & 1.6M  & 15h  & MoCap & VR Environment & SMPL-X & Action Label & \cmark &  \cmark \\
        LaserHuman~\cite{cong2024laserhuman} & 2024 & -           &  - & 11  &  3374 &   -   &  3h  & RGB + IMU & LiDAR & SMPL & Text & \cmark &  \cmark \\
        iReplica~\cite{guzov24ireplica}    & 2024 & 3DV           & 8  & 7   &   -   & 680K  & 0.8h &  IMU  & - & SMPL  & - &  \cmark &  \cmark \\
        ParaHome~\cite{kim2024parahome}    & 2024 & -             & 30 & 1   &  101  & 56M   & 7.3h & mRGB + IMU  & Structured Light & SMPL-X & Text &  \xmark &  \cmark \\
        CIRCLE~\cite{CIRCLE}               & 2023 & CVPR          & 5  & 9   &   7K  & 4.3M  & 10h  & MoCap & VR Environment & SMPL-X & - &  \xmark &  \xmark \\
        HUMANISE~\cite{humanise}           & 2022 & NeurIPS       & -  & 643 & 19.6K & 1.2M  &  -   & Alignment & mRGB-D & SMPL-X & Text &  \xmark &  \xmark \\
        GIMO~\cite{gimo}                   & 2022 & ECCV          & 11 &  19 & 217   & 129K  &  -   &  IMU  & Phone Scanned & SMPL-X & Gaze &  \cmark &  \xmark \\
        EgoBody~\cite{egobody}             & 2022 & ECCV          & 36 & 15  & 125   & 220K  &  -   &  mRGB-D & Phone Scanned & SMPL-X & Gaze &  \cmark &  \xmark \\
        RICH~\cite{RICH}                   & 2022 & CVPR          & 22 &  5  & 142   & 577K  &   -  &  mRGB & Laser Scanned & SMPL-X &  - &  \cmark&  \xmark  \\
        SAMP~\cite{hassan2021stochastic}   & 2021 & ICCV           & 1 &  7 &  - & 185K & 1.6h & MoCap & CAD Model & SMPL-X & -  &  \xmark &  \xmark          \\
        HPS~\cite{HPS}                     & 2021 & CVPR          & 7  &  8  &  -    & 300K  &  3h  &  IMU  & - &  SMPL  &  - &  \xmark &  \xmark  \\
        GTA-IM~\cite{caoHMP2020}           & 2020 & ECCV          & 50 & 49  & 119   & 1.0M  &   -  &  GameEngine & Game Environment & 3D skeleton & - &  \cmark &  \xmark \\
        PROX~\cite{PROX}                   & 2019 & ICCV          & 20 &  12 &  60   & 100K  &  -   &  RGB  & Structured Light & SMPL-X   & - & \cmark &  \xmark \\
        i3DB~\cite{iMapper}                & 2019 & TOG          & 1  & 15  &  -    &   -   &  -   &  RGB   & Optimization & 3D skeleton   & - & \cmark  &  \xmark \\
        PiGraphs~\cite{savva2016pigraphs}  & 2016 & TOG         & 5  &  30 & 63    & 100K  &  2h  & RGB-D & mRGB-D & 3D skeleton  & Text & \xmark &  \xmark \\
        
	\hline
	\end{tabular}
	}
	\end{threeparttable}
	\vspace{-5pt}
\end{table*}

\paragraph{Grasping Hand Datasets.}
\label{sec:ho-data-hand}
Hand-grasp datasets capture isolated instances of hand-object interaction, providing detailed annotations of grasp poses and contact points. These datasets are foundational for training and evaluating models on grasp synthesis and pose estimation. DexYCB\cite{chao2021dexycb} offers a comprehensive dataset with RGB-D images and 3D annotations of hand poses and object poses for 3D-printed objects from the YCB benchmark~\cite{calli2015benchmarking}, facilitating studies on accurate grasp detection and pose estimation. ObMan\cite{hasson19_obman} leverages synthetic data to simulate diverse hand-object interactions with MANO~\cite{MANO_2017} hand models and object meshes, emphasizing the generation of realistic static grasps in controlled environments. ContactDB\cite{Brahmbhatt_2019_CVPR} provides high-resolution annotations of contact areas on hand and object surfaces, offering a tactile perspective to grasp modeling and enabling research on contact dynamics. These datasets focus on precise hand-object alignment, capturing essential spatial relationships that are critical for static grasp analysis and synthesis.
Additionally, HO-3D\cite{hampali2020honnotate} is the first markerless dataset of color images with 3D annotations of the hand and object, designed for benchmarking 3D hand-object pose estimation.

\paragraph{Whole Body HOI Datasets.}
\label{sec:ho-data-wholebody}
Whole-body interaction datasets extend beyond hand-object interactions to capture the interplay between the entire human body and objects.
GRAB~\cite{taheri2020grab} provides detailed 3D motion data of whole-body grasps, including articulated hands, body poses, and even facial expressions, while interacting with 51 everyday objects. It captures both static grasps and dynamic interactions, such as object handovers and usage, offering a comprehensive resource for modeling full-body grasping actions and contact dynamics.
COUCH\cite{zhang2022couch} and CHAIRS\cite{jiang2023full} target whole-body seated interactions, such as sitting, leaning, and adjusting posture while interacting with different types of furniture. They focus on contact points and posture dynamics, and provide fine-grained annotations for realistic full-body interactions in indoor settings.
BEHAVE\cite{bhatnagar2022behave}, HODome\cite{zhang2023neuraldome}, InterCap\cite{huang2022intercap}, IMHD2\cite{zhao2024imhoi} capture full-body human-object interactions in natural environments using multi-view RGB-D recordings. They provide annotations for 3D human and object tracking, offering rich datasets for understanding naturalistic whole-body motion during interactions, such as carrying, pushing, and holding.

\paragraph{Articulated Object Interaction Datasets.}
\label{sec:ho-data-articulate}
ARCTIC~\cite{fan2023arctic} dataset is designed to study dexterous bimanual manipulation with articulated objects, addressing the complexities of handling objects with moving parts, such as laptops or scissors. 
It includes over $2.1$ million RGB frames from multiple synchronized views, with paired 3D hand and object meshes for high-fidelity capture of hand-object interactions.

\paragraph{Text-Conditioned HOI Datasets.}
\label{sec:ho-data-text}
Text-conditioned HOI datasets provide multimodal data linking natural language descriptions to human-object interactions.
HIMO\cite{lv2024himo} offers richly annotated sequences of full-body interactions with objects, paired with textual descriptions that capture detailed semantics and temporal progression, supporting tasks like text-to-motion synthesis and contextual interaction modeling. 
OMOMO\cite{li2023object} also aligns textual instructions with object-specific full-body interactions, focusing on generating diverse and realistic motions conditioned on language input and object properties. Semantic-HOI\cite{yang2024f} is built by prompting GPT-4V~\cite{achiam2023gpt} using hand-object images for annotations and includes fine-grained decoupled human pose descriptions.

\vspace{-6pt}
\subsection{Human-Scene Interaction Datasets}
\label{sec:hs-data}
Technological advances in motion capture, virtual reality, and multi-modal data acquisition have enabled the development of diverse HSI datasets. These datasets can be classified into two main categories: interactions with~\textit{static scene} and interactions with~\textit{dynamic-involved scenes}. Table \ref{table:hsmg_datasets} provides detailed statistics of current HSI datasets in these two categories.

\begin{table*}[t]
	\centering
	\captionsetup{font=small}
	\caption{\small{\textbf{Human-mix interaction datasets.} This table summarizes key statistics and features of various human-mix interaction datasets. Tasks: Types of human interaction tasks—HHI: Human-Human Interaction, HOI: Human-Object Interaction; Subjects: The number of entities involved in the dataset; Sequences: The number of motion clips available; Frames: The total number of frames capturing 3D human motions; Length: The cumulative duration of the dataset's motion data (in hours); Acquisition: The method used to obtain motion data; Modality: The representation format of motion data; Video, Text, Audio: Indicates whether the dataset includes corresponding modalities.}}
	\label{table:hmmg_datasets}
	\vspace{-5pt}
	\begin{threeparttable}
	\resizebox{0.99\textwidth}{!}{
	\setlength\tabcolsep{2.5pt}
	\renewcommand\arraystretch{1.02}
	\begin{tabular}{r||c|c|c|c|c|c|c|c|c|c|c|c}
	\hline\thickhline
	\rowcolor{mygray1}
        Dataset & Year & Venue & Tasks & Subjects & Sequences & Frames & Length & Acquisition & Modality & Video & Text & Audio \\
	\hline
	\hline
        HOI-M\(^3\)~\cite{zhang2024hoim3} & 2024 & CVPR & HHI, HOI & 46 Human, 90 Objects & - & 181M & 20h & MoCap & SMPL & \cmark & \xmark & \xmark \\
        HHOI~\cite{shu2016HHOI} & 2016 & IJCAI & HHI, HOI & 8 Human, 2 Objects & 118 & 7.5K & 0.2h & RGB-D & 3D Skeleton & \cmark & \xmark & \xmark \\
	\hline
	\end{tabular}
	}
	\end{threeparttable}
	\vspace{-5pt}
\end{table*}

\paragraph{Interaction with Static Scene.}
Static scene interaction datasets capture human behavior within fixed environments, focusing on interactions with stationary objects and unchanging scenes. These datasets vary in their capture methodologies and included modalities.
Early datasets, like PiGraphs~\cite{savva2016pigraphs}, establish foundations by capturing temporal interactions between humans and local object configurations. PROX~\cite{PROX} and i3DB~\cite{iMapper} advance the field by providing RGB video-optimized human motion data, while GTA-IM~\cite{caoHMP2020} introduces synthetic data collection through game engine interfaces, incorporating RGB images, pose visualizations, and depth maps. SAMP\cite{hassan2021stochastic} utilizes a pre-obtained CAD model and a motion capture device to obtain a diverse sitting scenario in the scene.
Recent datasets have expanded to include multiple sensory modalities. HPS~\cite{HPS} employs head-mounted cameras for subject self-localization in large 3D scenes, while RICH~\cite{RICH} provides vertex-level contact labels and multi-view videos for detailed human-scene contact analysis. EgoBody~\cite{egobody} and GIMO~\cite{gimo} incorporate eye gaze data alongside third-person and egocentric views, with GIMO specifically focusing on leveraging gaze information to predict future body movements.
HUMANISE~\cite{humanise} takes a novel approach by aligning pre-captured human motion sequences with scanned indoor scenes, additionally providing text descriptions of actions and interaction targets for multi-modal analysis. CIRCLE~\cite{CIRCLE} focuses specifically on reaching motions in cluttered environments, utilizing virtual reality (VR) to present scenes to motion capture subjects.

\paragraph{Interaction with Dynamic Scene.}
Human interactions inherently modify their surrounding environments, creating dynamic and evolving scenes. To capture these complex interactions, several datasets have been introduced that document scenarios involving moving objects, scene modifications, and interactions with articulated entities.
iReplica~\cite{guzov24ireplica} provides comprehensive motion capture data while tracking environmental changes induced by human actions, such as table displacement and door manipulation. Complementing this, ParaHome~\cite{kim2024parahome} specializes in fine-grained interactions with small objects, documenting both body and finger motions alongside dynamic changes in articulated objects—from laptop operations to stove control manipulation.
LaserHuman~\cite{cong2024laserhuman} offers a unique perspective by combining human motion capture with free-form language descriptions of movements and environmental interactions. This dataset encompasses both indoor and expansive outdoor environments, representing dynamic scenes through LiDAR point cloud technology. TRUMANS~\cite{TRUMANS} enhances the field by capturing diverse indoor activities with dynamic objects in a VR-assisted environment, providing frame-wise action labels for multi-modal analysis of activities such as bottle handling and refrigerator operation.
Building upon these foundations, LINGO~\cite{LINGO} stands as the most extensive human-scene interaction motion dataset to date. Utilizing VR-assisted capture techniques, it offers fully text-annotated documentation of diverse and extended human motions coupled with dynamic object interactions.

\vspace{-6pt}
\subsection{Human-Mix Interaction Datasets}
\label{sec:hm-data}
Compared to datasets focusing on primary interactive tasks, datasets capturing multi-entity interactions remain limited in scope. Currently, only two datasets—HOI-M\(^3\)\cite{zhang2024hoim3} and HHOI\cite{shu2016HHOI}—provide resources for studying human-human-object interactions. While HHOI is restricted to object root translation data, HOI-M\(^3\) offers comprehensive object mesh and motion information, enabling more sophisticated modeling of human-human-object interaction dynamics.

This scarcity presents significant opportunities for dataset development across various multi-entity scenarios, including human-human-object interactions, human-human-scene interactions, and complex human-human-object-scene combinations. A detailed comparison of existing human-mix interaction datasets is presented in Table~\ref{table:hmmg_datasets}.

\begin{table*}[t]
	\centering
	\captionsetup{font=small}
	\caption{\small{\textbf{Overview of evaluation metrics for human interaction motion generation.}}}
	\label{table:eval_metrics}
	\vspace{-5pt}
	\begin{threeparttable}
	\resizebox{0.99\textwidth}{!}{
	\setlength\tabcolsep{2.5pt}
	\renewcommand\arraystretch{1.4}
	\begin{tabular}{r||c|c}
	\hline\thickhline
	\rowcolor{mygray1}
        Category & Sub-category & Metrics \\
	\hline
	\hline
	\multirow{3}{*}{\textbf{Fidelity}} 
                & Comparison with Ground-Truth 
                & \makecell[l]{
                    MPJPE~\cite{ionescu2013human3}, MPJVE~\cite{ghosh2024remos}, Trajectory Error~\cite{guidedmotiondiffusion}, Location Error~\cite{guidedmotiondiffusion},
                    Average Error~\cite{guidedmotiondiffusion}, Root Error~\cite{shafir2023ComMDM}, \\
                    Pose Error~\cite{wang2021MRT}, AFD~\cite{goel2022interactionmixmatch}, AME~\cite{guo2022multipersonexmotion}, VIM~\cite{Adeli_2021_tripod}, VAM~\cite{Adeli_2021_tripod}, VSM~\cite{Adeli_2021_tripod}, NDMS~\cite{tanke2021intention}, SSCP~\cite{SocialDiffusion}, SJP~\cite{Baruah_2020_CVPRW}, PCK~\cite{ahuja2019reactornot}
                } \\ \cline{2-3}
                & Naturalness
                & \makecell[l]{
                    FID (FVD)~\cite{NIPS2017_FID}, Recognition Accuracy~\cite{guo2020action2motion}, MMD~\cite{maheshwari2021mugl}, Critic Accuracy~\cite{Kundu_2020_WACV}, Critic Score~\cite{motioncritic2025} \\
                } \\ \cline{2-3}
                & Physical Plausibility & Foot Skating Ratio~\cite{guidedmotiondiffusion} \\ \hline
	\multirow{1}{*}{\textbf{Diversity}} 
                & Inter-Motion & Diversity~\cite{guo2020action2motion}, Multimodality~\cite{guo2020action2motion} \\ \hline
        \multirow{5}{*}{\textbf{Condition Coherence}} 
                & Partners' Motion-Conditioned & Mutual Consistency~\cite{Tanaka_2023_ICCV}, BED~\cite{siyao2024duolando}, Contact Frequency~\cite{siyao2024duolando}\\ \cline{2-3}
                & Object-Conditioned & CD~\cite{bcd2021wu}, Contact Ratio~\cite{zuo2024graspdiff}, Penetration Depth/Volume/Percentage~\cite{zuo2024graspdiff}, Grasp Success Rate~\cite{braun2023physically} \\ \cline{2-3}
                & Scene-Conditioned & Collision Ratio~\cite{CIRCLE},  Penetration Distance~\cite{DIMOS}, Semantic Contact Score~\cite{zhao2022compositional}\\ \cline{2-3}
                & Text-Conditioned & R-Precision~\cite{HumanML3D}, MultiModal Distance~\cite{HumanML3D}, EID~\cite{ponce2024in2in}\\ \cline{2-3}
                & Audio-Conditioned & BAS~\cite{siyao2022bailando}, SSCA~\cite{joo2019cvpr}\\ \hline
        \multirow{1}{*}{\textbf{User Study}} 
                & User Study
                & \makecell[l]{
                    Preference~\cite{wang2024intercontrol}, Rating (Motion Quality~\cite{ghosh2024remos}, Reaction Plausibility~\cite{ghosh2024remos},
                    Realness~\cite{wang2021MRT})
                } \\ \hline
	\end{tabular}
        }
        \end{threeparttable}
        \vspace{-5pt}
\end{table*}

\section{Evaluation Metrics}
\label{sec:eval}
Human interaction motion generation requires comprehensive evaluation across multiple dimensions. Current metrics assess three key aspects:~\textit{Fidelity}—the quality and naturalness of generated motions;~\textit{Diversity}—the variety of generated outputs; and~\textit{Condition Coherence}—the adherence to interaction constraints and task objectives. A comprehensive examination of evaluation metrics is presented in Table~\ref{table:eval_metrics}.

\vspace{-6pt}
\subsection{Interaction Motion Fidelity}
\label{sec:eval-fidelity}

Evaluating the fidelity of generated human interaction motions is crucial to ensuring their accuracy, realism, and physical plausibility. This section summarizes evaluation metrics for fidelity in three key aspects:~\textit{Comparison with ground-truth motions}, assessing generated motion using distance-based metrics;~\textit{Naturalness}, evaluating perceptual and statistical similarities between generated and real motions; and~\textit{Physical plausibility}, measuring adherence to real-world physical constraints and interaction dynamics.

\vspace{-6pt}
\subsubsection{Comparison with Ground-Truth}
\label{sec:eval-fidelity-gt}

Fidelity in interactive human motion generation is predominantly evaluated by comparing generated motions with ground-truth motions using distance-based metrics, with Mean Per-Joint Positional Error (MPJPE)~\cite{ionescu2013human3} serving as the primary measure. MPJPE quantifies the average spatial discrepancy between each corresponding joint in the predicted and actual poses, with lower values indicating higher fidelity.
For a skeleton \textit{S} and a frame \textit{f}, MPJPE is calculated as
\begin{equation}
MPJPE(f, S) = \frac{1}{N_S} \sum_{i=1}^{N_S} \| m_{f, S}^{(f)}(i) - m_{gt, S}^{(f)}(i) \|_2 \quad,
\end{equation}
where \( N_S \) represents the number of joints in skeleton \( S \). When considering multiple frames, the overall error is obtained by averaging the MPJPE values across all frames.

In addition to MPJPE, other related metrics, such as Mean Per-Frame Per-Joint Velocity Error (MPJVE)~\cite{ghosh2024remos}, Aligned Mean Error (AME)~\cite{guo2022multipersonexmotion}, Trajectory Error~\cite{guidedmotiondiffusion}, Location Error~\cite{guidedmotiondiffusion}, Average (Root) Error~\cite{guidedmotiondiffusion}, Pose Error~\cite{wang2021MRT}, Average Frame Distance (AFD)~\cite{goel2022interactionmixmatch}, Visibility-Ignored Metric (VIM)~\cite{Adeli_2021_tripod}, and Visibility-Aware Metric (VAM)~\cite{Adeli_2021_tripod} are purposed as complementary metrics. These metrics similarly assess various aspects of distance, orientation, and trajectory between the generated and ground-truth motions, providing a comprehensive evaluation of location, pose, and movement accuracy under different conditions.

Beyond direct L2 distance computations, some evaluation metrics leverage more information to provide deeper insights. Visibility Score Metric (VSM)~\cite{Adeli_2021_tripod} assesses the precision of predicted visibility scores for all joints by measuring the average Intersection over Union (IoU) and F1 Scores of the joints, ensuring reliable visibility predictions in future frames. Normalized Directional Motion Similarity (NDMS)~\cite{tanke2021intention} evaluates the alignment of motion directions and the ratio of movement magnitudes between predicted and real motions. Symbolic Social Cues Protocol (SSCP)~\cite{SocialDiffusion} examines the accuracy of social interactions within the generated motions, such as appropriate gestures for activities like talking or listening, aligning them with symbolic interaction states observed in the ground-truth. Salient Joints Precision (SJP)~\cite{Baruah_2020_CVPRW}, measures the precision of predicted salient joints by comparing the generated sequence of important movement joints against the true sequence. Probability of Correct Keypoints (PCK)~\cite{ahuja2019reactornot} determines the accuracy of location keypoint predictions by verifying if they fall within a defined radius around ground-truth positions, providing a probabilistic measure of keypoint localization accuracy.

\vspace{-6pt}
\subsubsection{Naturalness}
\label{sec:eval-fidelity-naturalness}

Naturalness in human interaction motion generation assesses how lifelike and plausible the generated motions appear, often by comparing the statistical and perceptual properties of the generated motions to ground-truth motions.

Fréchet Inception Distance (FID)~\cite{NIPS2017_FID} and its video-adapted version, Fréchet Video Distance (FVD), have been widely used in many works~\cite{ghosh2024remos, xu2024regennet, ponce2024in2in, wang2024intercontrol, InterGen, shafir2023ComMDM, siyao2024duolando, xu2022actformer, Tanaka_2023_ICCV, Gupta_2023_DSAG, goel2022interactionmixmatch} to measure the divergence between the feature distributions of ground truth and generated motions. These metrics leverage deep features extracted from classification models to quantify the similarity between distributions. FID is calculated as:
\begin{equation}
\text{FID} = \| \mu_{gt} - \mu_{gen} \|_2^2 + \mathrm{tr} \left( \mathbf{C}_{gt} + \mathbf{C}_{gen} - 2 \left( \mathbf{C}_{gt} * \mathbf{C}_{gen} \right)^{\frac{1}{2}} \right),
\end{equation}
where \( \mu_{gt} \), \( \mu_{gen} \) and \( \mathbf{C}_{gt} \) and \( \mathbf{C}_{gen} \) are the means and covariance matrices of the deep features extracted from ground truth and the generated motions respectively. The operator \(\mathrm{tr}(\cdot)\) denotes the trace of the matrix.

Additionally, Maximum Mean Discrepancy (MMD)~\cite{maheshwari2021mugl} further assesses naturalness by evaluating the similarity between generated and real motion distributions either on a per-timestep basis (MMD-A) or across entire motion sequences (MMD-S) by flattening motion sequences into vector representations. Action Recognition Accuracy~\cite{guo2020action2motion} uses classification models to determine whether the generated motions can be accurately identified as specific actions, thereby measuring their realism and semantic correctness. Critic Accuracy~\cite{Kundu_2020_WACV} employs an adversarial model to assess how easily a discriminator can distinguish between real and generated motions; lower critic accuracy indicates higher naturalness. Critic Score~\cite{motioncritic2025} is a data-driven metric that learns human perceptual preferences to evaluate the naturalness and quality of generated motions, demonstrating higher consistency with human preference evaluations than other metrics.

\vspace{-6pt}
\subsubsection{Physical Plausibility}
\label{sec:eval-fidelity-pp}

Physical plausibility assesses how well the generated motions conform to realistic physical constraints and natural interaction dynamics. A common metric for this evaluation is the Foot Skating Ratio~\cite{guidedmotiondiffusion}, which quantifies unintended foot sliding during motion, reflecting the stability and grounding of movements.

\vspace{-6pt}
\subsection{Diversity}
\label{sec:eval-diversity}

In the evaluation of the diversity of generated motions, Diversity~\cite{guo2020action2motion} and Multimodality~\cite{guo2020action2motion} are critical metrics that assess different aspects of variation within the generated motions. Diversity refers to the variations among a set of generated motions, measuring the overall range and distinctness within the generated motion distribution. Diversity is calculated as:
\begin{equation}
\text{Diversity} = \frac{1}{S_d} \sum_{i=1}^{S_d} | v_i - v'_i |_2,
\end{equation}
where \( S_d \) is the number of samples used in the experiments, \( v_i \) and \( v'_i \) are the deep feature vectors of the \( i \)-th samples from two randomly sampled subsets of generated motions, and \( \| \cdot \|_2 \) represents the L2 norm (Euclidean distance).

In contrast, Multimodality measures how much the generated motions diversify within each action type or text prompt, reflecting the model’s ability to produce multiple plausible variations for a single input. Given a set of motions with \( C \) action types, for the \( c \)-th action, two subsets with the same sample size \( S_l \) are randomly sampled. The multimodality is formalized as:
\begin{equation}
\text{Multimodality} = \frac{1}{C \times S_l} \sum_{c=1}^{C} \sum_{i=1}^{S_l} \| v_{c,i} - v'_{c,i} \|_2,
\end{equation}
where \( v_{c,i} \) and \( v'_{c,i} \) are the deep feature vectors of the \( i \)-th samples from two randomly sampled subsets within the \( c \)-th action type.

In summary, Diversity ensures a wide range of motion variations, enriching the generated data, while Multimodality ensures the model can produce multiple plausible variations for a given input, enhancing its adaptability and usefulness in dynamic interactive scenarios.

\vspace{-6pt}
\subsection{Condition Coherence}
\label{sec:eval-cc}

Ensuring coherence in condition-driven human interaction motion generation is essential for producing coherent and contextually accurate interactions. This section evaluates condition coherence across multiple domains, including partners' motion, object, scene, text, and audio.

\vspace{-6pt}
\subsubsection{Partners' Motion-Conditioned}
\label{sec:eval-cc-human}

Partners' motion-conditioned coherence focuses on ensuring that the generated motions adhere closely to the partner's motions in terms of interaction coherence or rhythmic synchronization. One metric is Mutual Consistency~\cite{Tanaka_2023_ICCV}, which assesses the coherence of interactions between actor and reactor motions by examining factors such as relative positions, action-reaction timings, and directional alignment. This metric employs a trained classification model to distinguish between consistent (correct and unmodified interactions) and inconsistent (randomly paired) motion pairs. Another metric is Beat Echo Degree (BED)~\cite{siyao2024duolando}, which quantifies the consistency of dynamic rhythms between two dancers. BED is calculated by first identifying the timing of beats in both the leader and follower movements through the detection of local minima in motion velocity. Mathematically, BED is expressed as:
\begin{equation}
\text{BED} = \frac{1}{|B_l|} \sum_{t' \in B_l} \exp \left\{ -\frac{\min_{t_f \in B_f} \| t_l - t_f \|^2}{2\sigma^2} \right\},
\end{equation}
where \( B_l \) and \( B_f \) represent the sets of beat times for the leader and follower, respectively. \(\sigma\), the parameter used for normalization, is set to 3 in Duolando~\cite{siyao2024duolando}.

In terms of measuring physical plausibility in HHI, Contact Frequency~\cite{siyao2024duolando} can be used as a key metric that measures the frequency of physical interactions between individuals. Higher contact frequencies suggest more believable human-human interactions, as they align with the natural tendency for frequent contact in close interactions, such as dance or combat engagements.

\vspace{-6pt}
\subsubsection{Object-Conditioned}
\label{sec:eval-cc-object}
Evaluating human-object interaction requires metrics that rigorously assess both physical plausibility and the contact accuracy between humans and objects, particularly focusing on hand-object interactions where precise alignment is critical. Chamfer Distance~\cite{bcd2021wu} is widely used to measure the geometric consistency between predicted and ground-truth point clouds, making it effective in evaluating the accuracy of contact regions and the spatial alignment of hand poses with objects.
Contact ratio~\cite{zuo2024graspdiff, kwon2024graspdiffusion} directly measures the correctness of predicted contact points relative to ground-truth regions, reflecting how well the interaction captures realistic touch dynamics.
To ensure physical plausibility, several penetration-related metrics are employed.
Penetration Percentage ~\cite{paschalidis20243d} evaluates whether the signed distance between hand and object meshes exceeds a specified threshold (e.g. $1$ $mm$), identifying the proportion of mesh points involved in penetrative errors.
Penetration Volume~\cite{zuo2024graspdiff, Liu_2023_ICCV} quantifies the extent of penetration by voxelizing both hand and object meshes with $0.5$ $cm^3$ patches, counting the intersecting voxels.
Penetration Depth~\cite{paschalidis20243d, zuo2024graspdiff, liu2024geneoh} computes the
maximum distance from the intersected voxels to another
mesh surface. 

Additionally, Grasp Success Rate~\cite{braun2023physically} serves as a dynamic evaluation metric, where a grasp is considered successful if the object remains stable—without falling to the ground or table—within a 0.5 second time window after the interaction.
Another crucial metric is Simulation Displacement~\cite{zuo2024graspdiff}, which evaluates the stability of generated interactions by placing them in a physics simulation. By fixing the hand poses, this metric measures the displacement of the object's center of mass under the influence of gravity over a fixed period.

\vspace{-6pt}
\subsubsection{Scene-Conditioned}
\label{sec:eval-cc-scene}

Scene-conditioned coherence mainly focuses on evaluating the physical plausibility of human motion within a given scene. The Collision Ratio~\cite{CIRCLE} measures the frequency of body vertices or frames that penetrate the scene, highlighting the occurrence of physical rule violations. Penetration Distance~\cite{DIMOS} quantifies the extent of penetration, either as an average across all frames or as the maximum distance observed in a single frame. Together, these metrics provide insights into adherence to physical constraints in HSI scenarios.

Beyond physical rules, the Semantic Contact Score~\cite{zhao2022compositional} evaluates how well-generated interactions align with action-specific body-scene contact semantics. It is computed as a weighted sum of binary contact features, with weights derived from action-specific body vertex contact probabilities.

\vspace{-6pt}
\subsubsection{Text-Conditioned}
\label{sec:eval-cc-text}

Text-conditioned coherence metrics measure the alignment between generated motions and provided textual descriptions. R-Precision~\cite{HumanML3D} measures retrieval accuracy by ranking Euclidean distances between motion and text features within a dataset containing the ground truth and mismatched descriptions. It assesses precision at the top-1, top-2, and top-3 positions separately, determining how often the correct description is among the closest matches. MultiModal Distance~\cite{HumanML3D} calculates the average Euclidean distance between the motion features of the generated motions and the text features of their corresponding descriptions, providing a direct measure of the feature space alignment between modalities. Additionally, Extrinsic Individual Diversity (EID)~\cite{ponce2024in2in} measures how distinct textual descriptions influence the diversity of individual motion dynamics by comparing distributions of generated motions generated with original and randomly replaced descriptions via Wasserstein distance. A higher EID indicates greater control over the diversity of motions driven by individual textual inputs. When analyzed alongside other metrics like R-Precision and FID, EID helps assess the balance between individual diversity and interaction quality, ensuring that the generated motions are both varied and accurately conditioned on the provided text descriptions.

\vspace{-6pt}
\subsubsection{Audio-Conditioned}
\label{sec:eval-cc-audio}

Audio-conditioned coherence emphasizes the alignment of generated motions with corresponding audio inputs, such as music rhythms or speech patterns. A key metric used to evaluate this alignment is the Beat-Align Score (BAS)~\cite{siyao2022bailando}, which assesses how well the generated motions synchronize with the rhythmic beats of the music. When the reference base is replaced with the leader's dynamic beats, BAS is adapted to Beat Echo Degree (BED), as described partner's motion-condition coherence. Additionally, Speaking Status Classification Accuracy (SSCA)~\cite{joo2019cvpr} measures the model's ability to generate contextually appropriate motions that reflect the speaking status of the audio input. This metric involves classifying whether the generated motion corresponds to speaking or non-speaking states, ensuring that the motions are contextually appropriate and effectively represent audio-driven activities such as gestures associated with speech. These metrics provide diverse perspectives for assessing the model's ability to generate contextually accurate and interactive human motions driven by audio input.

\vspace{-6pt}
\subsection{User Study}
\label{sec:eval-user-study}

User studies play a crucial role in evaluating interactive human motion generation, particularly for assessing objectives that are challenging to quantify through automated metrics alone. These studies provide valuable insight into the subjective quality and perceptual aspects of generated motions. User studies can be broadly categorized into two major types. The first type involves eliciting user preferences~\cite{wang2024intercontrol} by comparing generated motions against baseline methods or ground-truth motions, allowing researchers to determine which motions are favored in terms of overall appeal and effectiveness. The second type requires participants to rate the generated motions on various dimensions, such as motion quality~\cite{ghosh2024remos}, reaction plausibility~\cite{ghosh2024remos}, and realism~\cite{wang2021MRT}. These ratings help in evaluating specific attributes of the generated motions, providing a nuanced understanding of their strengths and areas for improvement.

\section{Conclusion and Outlook}
\label{sec:conclusion}

In conclusion, this survey provides a systematic review of recent advances in human interaction motion generation, organized around four interaction categories: human-human, human-object, human-scene, and human-mix interactions. For each category, we examine methodological breakthroughs and classify approaches based on their solutions to core technical challenges. We analyze the distinctive characteristics of relevant datasets and their contributions to the advancement of the field. In addition, we present a unified framework for evaluating different interaction tasks through standardized metrics. Given the rapid evolution of this field, we identify and discuss four promising research directions that warrant further investigation.

\paragraph{Data.} Acquiring high-quality human interaction data presents significant challenges and costs. For instance, capturing human-scene interactions requires either comprehensive scene scanning or the meticulous reconstruction of virtual 3D scene layouts in physical spaces, followed by motion capture and coordinate system calibration~\cite{TRUMANS,LINGO}. While recent datasets have documented diverse human interactions in controlled environments, they struggle to encompass the full spectrum of real-world interaction scenarios. 
Several promising approaches could address these limitations. First, more efficient motion capture technologies, such as IMU sensors, could provide a trade-off between data quality and coverage. Second, leveraging heterogeneous data sources could enhance motion modeling; for example, generative motion priors could be learned from large-scale datasets of isolated motions. These motion priors can be further applied to interacting with other entities. Additionally, LLMs and VLMs, which encode rich knowledge of human interactions, could help overcome the data scarcity challenge while improving flexibility and generalizability.

\paragraph{Physical Plausibility.} Current approaches for generating interactive motions face challenges in achieving physical plausibility, such as accurately representing gravity, force interactions, and texture responses. Physics simulator-based methods~\cite{unihsi} offer a partial solution to these challenges; they typically rely on reinforcement learning frameworks that use physical feedback as rewards for motion correction. However, these approaches have limitations: they are often constrained to specific actions and struggle to generalize to novel motions. Furthermore, they exhibit limited compatibility with expressive generative models such as diffusion models. Future research directions may investigate hybrid approaches that combine the physical accuracy of simulator-based methods with the flexibility and generative capabilities of modern deep learning architectures.

\paragraph{Representation.} Efficient data representation of motions, 3D objects, and scenes plays a crucial role in interaction motion generation. An effective representation framework for interaction scenarios must capture both intrinsic properties (e.g. motion joint positions, object geometry) and relational information between interacting entities. In human-human interactions, Sebastian et al.\cite{starke2020local,starke2021neural} demonstrate this by extracting extensive motion features of characters relative to their opponents to characterize interaction dependencies. Similarly, for human-object interactions, InterDiff\cite{xu2023interdiff} reveals that object motions relative to contact points, rather than world origins, exhibit more learnable patterns.
The representation of 3D objects and scenes presents additional challenges due to their inherent complexity: they comprise thousands of unordered points and faces that lack well-defined structural representation. This challenge is particularly pronounced in human-scene interactions, where the scale of 3D geometric elements becomes intractable, compounded by the need to represent spatial layouts and regional semantic information. Given the limited availability of high-quality human interaction data, developing efficient, interaction-aware feature representations becomes paramount.

\paragraph{Editing and Controllability.} The ability to edit interactions and precisely control their generation is crucial for practical applications. While several techniques have been developed for single-person motion generation—including joint trajectory-based control~\cite{xie2023omnicontrol,pinyoanuntapong2024controlmm}, stylized generation~\cite{zhong2024smoodi}, and text-prompted editing~\cite{athanasiou2024motionfix}—these approaches have yet to be effectively adapted for human interaction scenarios. We believe these investigations could bring our interaction generation researches closer to real-world applications such as animations and robotics.

\section*{Data Availability Statement}

This study is a survey and does not involve the generation or analysis of new datasets. All referenced datasets are available from their original sources. The details of all referred works are also provided in this public GitHub repository: \url{https://github.com/soraproducer/Awesome-Human-Interaction-Motion-Generation}.

%
%

\bibliographystyle{spmpsci}      

\bibliography{
bib/intro,
bib/scope,
bib/preliminaries,
bib/interactive_hhmg,
bib/interactive_homg,
bib/interactive_hsmg,
bib/interactive_hmmg,
bib/popular_refs,
bib/future_works
}  


\end{document}